%% file: main.tex
\documentclass[conference]{IEEEtran}

\usepackage[final]{changes}

\usepackage[font=small,labelfont=bf]{caption}

\usepackage{lipsum}
\usepackage[mathlines,switch]{lineno}

\usepackage{diagbox}
\usepackage{threeparttable}
\usepackage{enumitem}
\usepackage{url}
\usepackage{graphicx}
\usepackage[vlined,ruled]{algorithm2e}
\usepackage{amsmath}
\usepackage{url}
\usepackage{subfigure}  
\usepackage{algorithmic}
\usepackage{tabu}
\usepackage{float} 
\usepackage{framed}
\usepackage{gensymb} 
\usepackage{array}
\usepackage{comment}
\usepackage{xcolor}
\usepackage{listings}
\usepackage{color}
\usepackage{fancyhdr}
\usepackage{gensymb}
\fancypagestyle{firstpage}
{
    \fancyhead[L]{}    
    \fancyhead[C]{\emph{To Appear at the 51st Annual IEEE/IFIP International Conference on Dependable Systems and Networks (DSN'21)}}
}

\definecolor{dkgreen}{rgb}{0,0.6,0}
\definecolor{gray}{rgb}{0.5,0.5,0.5}
\definecolor{mauve}{rgb}{0.58,0,0.82}

\begin{document} 

\long\def\authornote#1{%
  \leavevmode\unskip\raisebox{-3.5pt}{\rlap{$\scriptstyle\diamond$}}%
  \marginpar{\raggedright\hbadness=10000
    \def\baselinestretch{0.8}\tiny
    \it #1\par}}
\newcommand{\justin}[1]{\authornote{Justin: #1}}
\newcommand{\zitao}[1]{\authornote{Zitao: #1}}
\newcommand{\karthik}[1]{\authornote{Karthik: #1}}

\newcommand{\sysname}{\emph{Ranger}\xspace}

\author{
    \IEEEauthorblockN{Zitao Chen\IEEEauthorrefmark{1}, Guanpeng Li\IEEEauthorrefmark{2}, Karthik Pattabiraman\IEEEauthorrefmark{1}}
    \IEEEauthorblockA{\IEEEauthorrefmark{1}University of British Columbia \IEEEauthorrefmark{2}University of Iowa}
    \{zitaoc, karthikp\}@ece.ubc.ca \{guanpeng-li\}@uiowa.edu
}

\title{A Low-cost Fault Corrector for Deep Neural Networks through Range Restriction}

\maketitle
\thispagestyle{plain}
\pagestyle{plain}
\thispagestyle{firstpage}

\begin{abstract}
\input{sections/abstract}

\end{abstract}

\begin{IEEEkeywords}
Resilience, Machine Learning, Fault Correction
\end{IEEEkeywords}
\section{Introduction}
\input{sections/introduction}

\section{Background and Fault Model}
\input{sections/background}

\section{Methodology}
\input{sections/methodology}

\section{Implementation}
\input{sections/implementation}

\section{Evaluation}
\input{sections/evaluation}

\section{Discussion}
\input{sections/discussion}

\section{Related Work}
\input{sections/relatedwork}

\section{Conclusion} 
\input{sections/conclusion}

\section*{Acknowledgement}
This work was funded by a grant from the Natural Sciences and Engineering Research Council of Canada (NSERC), and grants from Huawei and Intel. We thank the anonymous reviewers of DSN'21 for their comments, which helped improve the paper, and our shepherd Prof. Evgenia Smirni.

\bibliographystyle{IEEEtran}
\bibliography{IEEEabrv,lib}

\end{document}

%% file: sections/abstract.tex
The adoption of deep neural networks (DNNs) in safety-critical domains has engendered serious reliability concerns. A prominent example is hardware transient faults that are growing in frequency due to the progressive technology scaling, and can lead to failures in DNNs.
This work proposes \sysname, a low-cost fault corrector, which \emph{directly rectifies the faulty output due to transient faults without re-computation}.
DNNs are inherently resilient to \emph{benign faults} (which will not cause output corruption), but not to  
\emph{critical faults} (which can result in erroneous output). \sysname is an \emph{automated transformation} to selectively restrict the value ranges in DNNs, which reduces the large deviations caused by critical faults and transforms them to benign faults that can be tolerated by the inherent resilience of the DNNs.
Our evaluation on 8 DNNs demonstrates \sysname significantly increases the error resilience of the DNNs (by 3x to 50x), with no loss in accuracy, and with negligible overheads.

%% file: sections/introduction.tex
\label{sec:intro}

In the recent past, there has been a wide adoption of deep neural networks (DNNs) in various domains~\cite{brown2020language,he2016deep,krizhevsky2012imagenet}. Many of these applications are also safety-critical~\cite{julian2016policy,xiong2017robust,rajpurkar2017cardiologist,esteva2017dermatologist,bojarski2016end}. One example is autonomous robots, in which DNNs are used for scene understanding and path navigation (e.g., aircraft collision avoidance~\cite{julian2016policy}). Another emerging example is Autonomous Vehicles (AVs), in which DNNs are used to provide end-to-end autonomy. DNNs are crucial for the AV system to perceive its surroundings (e.g., traffic sign detection) in order to safely maneuver the vehicle. Therefore, the safety and reliability of DNNs is becoming a critical consideration. 

On the other hand, modern computing systems are becoming increasingly susceptible to hardware transient faults (i.e., soft errors), which often arise due to high-energy particle strikes (e.g., terrestrial neutrons), transistor variability, and  thermal cycling. Transient faults typically manifest as \emph{single bit-flips} in the system, and they are predicted to become more frequent due to the effects of technology scaling~\cite{snir2014addressing,oliveira2017experimental,gomez2014gpgpus}. 
Prior studies have shown that transient faults can effectively result in failures of DNNs, leading to safety violations such as causing the AV system to miss obstacles in its path~\cite{li2017understanding,chen2019}.  {We refer to these failures as Silent Data Corruptions (SDCs).}


Traditional methods to protect systems from soft errors use redundancy at the hardware level (e.g., Triple-Modular Redundancy). However, such techniques are prohibitively expensive, thus making them difficult to deploy in practice. For example, the cost-per-unit and computation performance are of significant importance in many areas (e.g., industrial robotics, AVs). Duplicating hardware components adds to the total cost of the system, and requires synchronization and voting.  


 {
There have been many techniques proposed to enhance the error resilience of DNNs~\cite{li2017understanding,schorn2018efficient,hoang2019ft,mahmoud2020hardnn,zhao2020algorithm,ozen2019sanity}. Li \emph{et al.}~\cite{li2017understanding} uses unusual values as symptoms for fault detection~\cite{li2017understanding}. 
Hong \emph{et al.} proposes to mitigate memory faults by modifying the network's architecture (e.g., using Tanh as activation function)~\cite{hong2019terminal}.
Mahmoud \emph{et al.}~\cite{mahmoud2020hardnn} proposes to detect faults based on the mismatch between two duplicated computations. 
	However, these techniques often suffer from limited fault coverage~\cite{schorn2018efficient,mahmoud2020hardnn,hong2019terminal} and/or incur significant implementation effort and overhead~\cite{li2017understanding,schorn2018efficient,mahmoud2020hardnn} (more details in Section~\ref{sec:related-work}). Moreover, in these systems, program re-execution is often needed in order to recover the correct output. Re-execution incurs considerable overhead and is undesirable for time-critical systems such as AVs that have real-time constraints~\cite{AV-dataGenRate1,av-fitRate}.
	Therefore, there is a compelling need for an efficient technique to protect DNNs from transient faults without incurring significant overheads.
}

 {
To overcome the above challenges,
this work proposes \sysname \footnote{Similar to how a park ranger protects parklands, \sysname protects DNNs  from soft errors by restricting the ranges of values.}, a \emph{low-cost} fault correction technique that can \emph{directly rectify the faulty output} due to transient faults without any re-computation.} 
We leverage the insight that DNNs are inherently tolerant of \emph{benign faults} (i.e., faults that will not corrupt the program output in a significant way), and propose \sysname to transform the \emph{critical faults} (these are the faults that will result in SDCs) into benign faults. 
In particular, \sysname selectively restricts the ranges of values in specific DNN layers, thereby enabling the DNNs to generate correct output \emph{even in the presence of faults (in most cases).} 

 {
The intuition behind \sysname is that DNNs are statistical entities, and \emph{do not} always require exactness in their computations (e.g., incorrect computation of one internal neuron is acceptable 
if the final predicted label is correct). Thus,  range restriction is able to mitigate the large deviation caused by critical faults, so that the reduced deviation can be tolerated by the inherent resilience of DNNs. This eliminates the need to use expensive re-computation to derive the correct results. 
}

 {
We implement \sysname as an {\em automated technique} for the TensorFlow machine learning (ML) framework~\cite{abadi2016tensorflow}, which is the most widely used ML framework today~\cite{tf-popularity}.
\sysname \emph{automatically} converts the unreliable DNNs into resilient DNNs, without requiring any programmer intervention. }

\begin{figure}[t]
\centering
  \includegraphics[width=3.5in, height=1.2in]{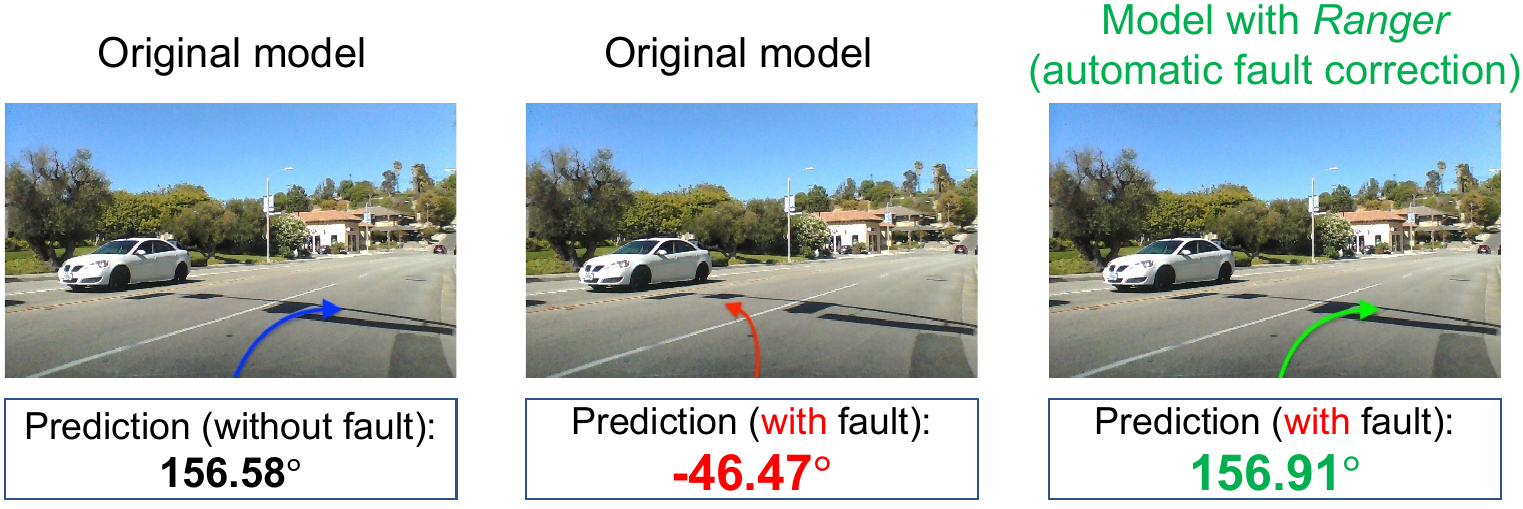}
\caption{An example to illustrate how \sysname enables automatic fault correction on a DNN deployed in Autonomous Vehicles (AVs).}
\label{fig:exp-self-driving}
\end{figure}

 {
Fig.\ref{fig:exp-self-driving} illustrates how a DNN augmented with \sysname can steer an AV in the presence of a transient fault. This is a real example for our experiments. As can be seen, the AV's steering angle changes from $ 156.58 \degree$ to $ -46.47 \degree $ due to a transient fault in the DNN inference. However, \sysname performs error correction to restore the faulty value to $156.91 \degree$, which though not exact, is sufficiently close to the original value to navigate the AV safely despite the fault. }

Prior work uses value truncation for different purposes in machine learning (ML) domain, such as improving the performance of DNNs~\cite{goodfellow2016deep,langford2009sparse}; robustness of DNNs to outliers~\cite{wu2007robust}, and privacy preservation~\cite{ji2014differential,shokri2015privacy} (Section~\ref{sec:related-work}). In contrast, we leverage value restriction for protecting DNNs from transient faults. 
\sysname is based on the idea of our prior work BinFI~\cite{chen2019}, which distinguishes between benign and critical faults based on the mathematical properties of DNN components, for speeding up fault injections. However, BinFI does  not propose any technique to protect the DNNs from the critical faults.
{\em To the best of our knowledge, \sysname is the first low-cost technique to provide efficient fault correction for transient faults in DNNs without requiring  re-computation.
}


The main contributions of this paper are as follows:

\begin{itemize}[leftmargin=*]
\item  {Propose \sysname, a technique to selectively restrict the ranges of values in DNNs, which can mitigate the large deviations by critical faults to those that can be tolerated by DNNs. \sysname is thus able to rectify the faulty output due to transient faults without performing  any re-computation.
} 
\item Present an approach to choose particular layers in DNNs for range restriction based on their value dependency, and derive the restriction bounds to be used by \sysname.  
\item \added{Implement \sysname in the  TensorFlow framework, to automatically make the DNNs resilient to transient faults,}
\item Perform an experimental evaluation of \sysname on 8 DNN models with a total of 5 datasets (including two DNN applications in the AV domain). The evaluation demonstrates that \sysname: 1) significantly enhances the resilience of the DNNs models - it reduces the SDC rates from $14.92\%$ to $0.44\%$ (in classifier DNNs), and from $23.76\%$ to $2.49\%$ (in the AV DNNs); 2) does not degrade the accuracy of any of the evaluated models,  
and 3) incurs negligible memory overheads, and $0.530\%$ performance overhead (on average). 
\end{itemize}

%% file: sections/background.tex
\label{sec:background}
In this section, we first provide an overview of deep learning, then describe the reliability challenge for safety-critical DNNs applications like AVs, followed by the fault model.

\subsection{Deep Learning}
\label{sec:background-deep-learning}

Deep learning is a field of artificial intelligence that typically leverages DNNs to address problems in both classification and regression.
In this paper, we primarily consider convolutional neural networks (CNNs), a class of DNNs application (others such as recurrent neural networks~\cite{goodfellow2016deep} are not considered).
A typical DNN uses multiple layers to progressively extract high level features from the raw input. The primary computation usually occurs in the convolution (Conv) layer that extracts the underlying patterns. The results are then fed to the activation (ACT) layer in which the ACT function is used to determine how the neuron should be activated (e.g., performing non-linear transformation). Other layers such as pooling, normalization layers can be added following the ACT layer. The layer closest to the output is usually called fully-connected (FC) layer. 
An DNN model typically goes through 2 phases: 1) training phase in which the model is trained to learn a particular task or set of tasks; 2) inference phase where the model is used for actual deployment. 
\subsection{ {Reliability of AVs}}
Autonomous vehicles (AVs) are systems that combine a variety of sensors (e.g., LiDAR) to perceive the surroundings and a control systems (e.g., DNNs) that interpret the sensory information to safely navigate the vehicles such as identifying obstacles in the path. Given the complexity of the road conditions, AVs usually have following requirements: 1)  process the humongous sensory information~\cite{AV-dataGenRate1,AV-dataGenRate2} - high throughput; 2) perform real-time action such as applying the brake before collision happens (e.g., an inference rate of 33 {ms}/frame to classify the frame images)~\cite{AV-dataGenRate1,av-fitRate} - low latency. 
These requirements present significant challenges for AV reliability.

Generally, faults in AVs that can result in safety violation (e.g., as per the ISO 26262 standard~\cite{iso26262}) can be categorized into systematic faults and transient faults~\cite{av-fitRate}. Systematic faults are induced by design defects in hardware and software, and they can be mitigated at design time~\cite{pei2017deepxplore,tian2018deeptest,rubaiyat2018experimental}. Transient faults, however, need runtime mitigation, which incurs performance overheads. 
Further, many of these applications have to adhere to standards, e.g., the ISO 26262 standard~\cite{iso26262} for AVs requires that there should be no more than 10 FIT (Failure in Time), which is 10 failures in a billion hours of operation. 
Therefore, it is important to design efficient mitigation techniques for soft-errors that can cause SDCs.

 {
To take one example, a recent study \cite{av-fitRate} has estimated that an AV needs to classify each frame image in 33 \emph{ms} interval to safely navigate the vehicle. YoLo V5, a state-of-art object detection model released in May 2020, needs 25.5ms for each inference~\cite{yolo}, which means the fault protection budget should be lower than 30\%; otherwise it could lead to slowdown in response and may result in reaction-time based accidents~\cite{banerjee2018hands}. Note that this is a conservative estimate as there is variation from frame to frame, and as a safety margin needs to be built in. Therefore, the fault detection and correction should have very low overhead, while having high coverage for SDCs.}

\subsection{Fault Model}
\label{sec:fault-model}

In this study, we consider hardware transient faults (i.e., soft errors) that occur randomly during the inference phase of the DNNs {(the input to the program is correct)}. We consider the inference phase because DNNs are usually trained once, while the inference tasks are performed repeatedly in deployment, and are hence much more likely to experience faults during their lifetime. 
Therefore, faults in the inference phase can have severe consequences (e.g., misclassify the ``stop'' sign as ``yield'' sign in an AV)
~\cite{reagen2018ares,li2017understanding,schorn2018efficient,chen2019}. 
	
We assume that faults can only arise in the processor's data path (ALUs and pipeline registers); and that the main memory, cache and register file are protected by ECC or parity~\cite{NIPS2019_8810}. Hence, we do not consider faults in these components.  
This is in line with previous reliability studies~\cite{li2018modeling, ashraf2015understanding, georgakoudis2017refine}.
In addition, we assume that faults do not arise in the control logic of the processor, as that constitutes only a small fraction of the total area of the processor. We only consider activated faults as masked faults do not affect the program's execution.   


We also assume that at most one fault occurs per program execution, because soft errors are relatively rare events given the typical inference time of DNNs. This also follows the fault models in prior studies~\cite{li2018modeling,ashraf2015understanding,fang2014gpu,wei2014quantifying,chen2019}.

 Finally, we inject single bit flips in the software implementation of the DNN, as a transient fault is often manifested as a single bit flip at the software level~\cite{Chang2018}, and studies have shown that multiple bit-flip errors result in similar fault propagation patterns as single bit-flip errors that cause SDCs~\cite{sangchoolie2017one, chang2018evaluating}. With the above said, we also evaluate our work under the multi-bit flip fault model in Section~\ref{sec:multi-bit-ranger}.

%% file: sections/methodology.tex
\label{sec:methodology}

To protect the DNNs from transient faults (particularly critical faults), we propose to  selectively restrict the ranges of values in different DNN layers. This technique that we call \sysname, is based on two properties in DNNs: 1) the monotonicity of operators in DNNs~\cite{chen2019}; 2) the inherent resilience of DNNs to insignificant errors~\cite{venkataramani2014axnn}. 

In this section, we first discuss how transient faults result in output corruption in DNNs (e.g., misclassification). 
We then explain how the characteristics of critical faults in DNNs, and the inherent resilience of DNNs can be leveraged to improve the error resilience of DNNs.
Finally, we elaborate on the design of \sysname to enable effective fault correction.
 
\subsection{Fault Propagation in DNNs}
Unlike traditional programs in which soft errors can corrupt the outputs by diverting the control flow~\cite{li2018modeling,hari2012relyzer}, faults that occur in DNNs are typically numerical deviations (e.g., change of the activation value in the feature map). Even the faults that modify connections between neurons can be modeled as numerical deviations. 
For example, a fault that causes the DNNs to break the connection between two neurons can be considered as a 0 value passed between them. 

The output quality of the DNNs is determined by the numerical output. For classification tasks, the output for the correct label needs to be high, and for the incorrect labels, it needs to be low. For regression tasks, the output needs to be as close to the target variable as possible. Therefore, for a fault to corrupt the output of DNNs, it needs to inflict significantly large deviation at the output. 

\begin{figure}[t]
\centering
  \includegraphics[width=3.5in, height=1.7in]{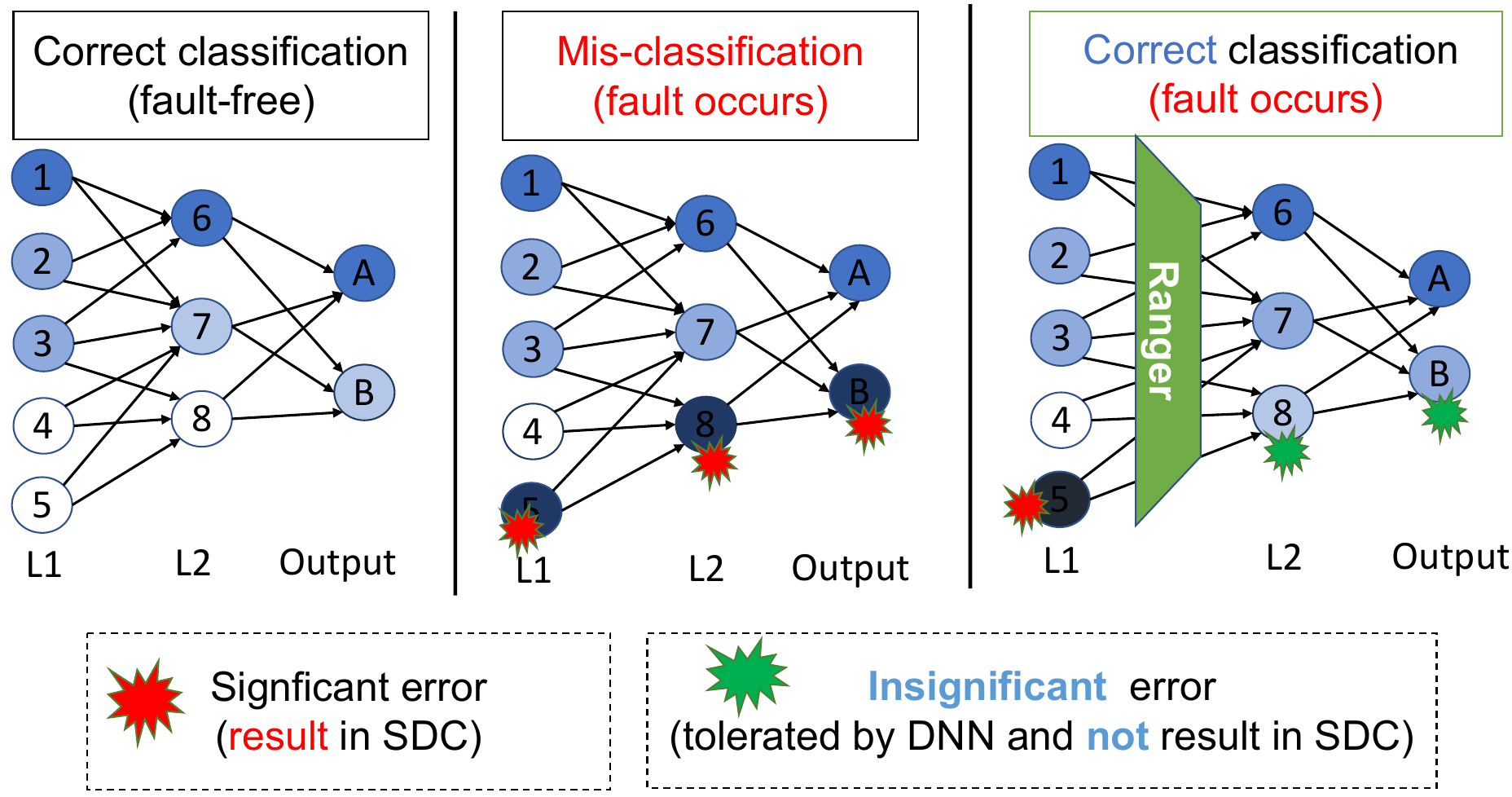}
\caption{Example of a fault causing misclassification (SDC), and how \sysname enables fault correction by dampening the error to be tolerated by the DNN. 
Darker colors represent larger values. We assume that label A is the correct label for the DNN under fault-free execution.}
\label{fig:fault-prop-exp}
\end{figure}

We provide an example on how a fault could effectively result in a misclassification in an DNN, in Fig.~\ref{fig:fault-prop-exp}. For simplicity, the figure only shows two hidden layers L1 and L2, where the colored nodes denote the activated neurons. The darker nodes are those with larger activation values (e.g., Node 6 has a larger activation value than that of Node 7). 
Though the fault is propagated through all the subsequent connected neurons, not all of them will result in large deviations - smaller deviations are not highlighted for simplicity (e.g., fault propagating from Node 5 to Node 7 result in small deviations because  the weight connection from Node 5 to Node 7 is small). 

In the fault-free execution, the classification result is assumed to be correct, i.e., label A. However, when a fault occurs during the computation (Node 5 in the center network in Fig.~\ref{fig:fault-prop-exp}), it causes a large deviation in the faulty neuron, and this subsequently results in the output for label B to be higher than that of label A. This fault thus leads to a misclassification, and is hence a critical fault. {\em We consider any DNN output that deviates from the correct output of the program as an SDC, e.g., an image misclassification.}

\subsection{Intuition behind Range Restriction}
\label{sec:exploit-inherent-resilience}

%

\textbf{Monotonicity}. In prior work, we study the mathematical properties of DNNs, and find that many of the computations in DNNs (e.g., ReLu, SoftMax, multiply-accumulate (MAC)) exhibit the monotone property~\cite{chen2019}: \textit{monotonically increasing}: $f(x_i) \geq f(x_j), \forall x_i > x_j$; or \textit{monotonically decreasing}: $f(x_i) \leq f(x_j), \forall x_i > x_j$. 
For example, assume a fault occurs at the Multiply application in the MAC operation and two faults $x_1, x_2$ occur at different bits - $x_1$ is at the high-order bit. The monotone property is satisfied as: $|x_1w_i| \geq |x_2w_i|$, where $w_i$ is the weight. Therefore, the fault propagation effect caused by $x_1$ is larger than that by $x_2$ and thus is \emph{more likely} to result in an SDC.  
{This property has been found in the common operations (e.g., Convolution, ReLu, etc.) employed in common DNN architectures (e.g., VGGNet, ResNet)~\cite{chen2019}.} 

Thus, the fault propagation behavior can be approximated as a monotone function.  
This monotone property implies that for faults to cause large deviations at the output layer, they should also cause large deviations at the place where they occur.  
This leads to the observation that critical faults tend to cluster around high-order bits (i.e., causing large value deviation), while faults at low-order bits tend to be masked and do not typically affect the output much. 
Thus, range restriction is analogous to ``transferring'' the faults from the high-order bits to the low-order bits (since the effect caused by fault is reduced), which can be tolerated by the model itself.

\textbf{Mitigating critical faults.} Typically critical faults in DNNs corrupt the output by fault amplification, which would propagate a single fault into multiple values causing large value deviations. Restricting the values in the DNNs reduces the deviations caused by the faults, which in turn lowers the chance of the fault to result in an SDC. 
This is because DNNs inherently can tolerate small value deviations and generate correct outputs regardless. \sysname leverages this property to prevent critical faults from corrupting the outputs, while tolerating other faults. We provide an example in Fig.~\ref{fig:fault-prop-exp}, where the goal of \sysname is to reduce the deviation by the critical faults (i.e., significant error) into smaller ones (i.e., insignificant error). Despite the insignificant errors, the system is able to generate the correct output due to the inherent resilience of DNNs. This property enables effective fault correction by \sysname, without the need to re-compute the result.

\textbf{Maintaining the accuracy of original models.} 
{While \sysname is intended for fault correction, it is also important to ensure that it does not degrade the accuracy of the original models (in the absence of faults). 
Because \sysname derives the restriction bound from the training data (see next section), it is possible for it to truncate some naturally large values in the DNNs. This is especially so when the DNN is tested on unseen data (we use a separate \emph{validation set} that is different from training set to simulate unseen data in our evaluation). 
We find that though this can occur, it rarely happens in practice as the restriction bound collected from the training data is sufficiently representative of the value ranges (e.g., in the VGG16 model, only in 5 out of 50,000 cases, do the values exceed the restriction bound when the network is tested on unseen data, in the absence of faults). 
Moreover, even in the above scenario, \sysname is still unlikely to degrade the accuracy of the model. This is because \sysname only truncates the values to the restriction bound. Such value reduction is often acceptable and can be tolerated inherently by the DNNs, without degrading its accuracy (Section~\ref{sec:results}).
}
Note that we do not require the training data to represent the faulty outcomes (unlike in prior work~\cite{schorn2018efficient,mahmoud2020hardnn}), and hence we do not need to perform expensive fault injection (FI) experiments for \sysname. 

\subsection{Selective Range Restriction}
\label{sec:range-rest}
In this section, we explain how to implement \sysname in DNNs. 
The first step is to {\em derive the restriction bounds} from the network through profiling, after which the network is transformed to {\em insert Ranger} on selected layers. Finally, the protected DNN with \sysname is released for deployment.

\textbf{Step 1: Deriving restriction bounds.}
The restriction bound can be derived from: 1) the function itself, e.g., Tanh function has an inherent bound of $(-1,1)$: For these functions, we do not have to derive the restriction bounds from the ACT value distribution; 2) Statistical sampling: For functions that are unbounded (such as the ReLu function that does not have an upper bound), the restriction bound can be derived from sampling the distribution of the values in the function, from which we can choose an appropriate bound. 

The selection of the bound can be adjusted based on whether we are willing to accept accuracy loss for resilience boosting. 
A conservative approach is to set the bound to the maximal value such that it is less likely to affect the accuracy of the model. Alternatively, we can also choose a smaller bound to gain higher resilience boosting at the cost of accuracy. We choose the conservative approach as the default configuration, but also evaluate the latter approach (Section~\ref{sec:discussion}).

\textbf{Step 2: Inserting Ranger into selected DNN layers.}
After deriving the restriction bound, the next step is to apply \sysname into the selected layers in DNNs. 
A DNN typically consists of many different layers~\cite{goodfellow2016deep}: Convolution, Activation (ACT), Pooling, Normalization, Fully-connected layers, etc. While these different layers can be considered for range restriction, we find that range restriction on the ACT layer is particularly desirable for two reasons: 
1) ACT function is used to determine the response from the neuron's output (e.g., filtering out the negative output in ReLu ACT function).  This particular feature of ACT also makes it ideal to be used for ``filtering'' out the potential critical faults. For example, an ACT function can restrict a large output from the previous layer (where a transient fault occurs). This restriction 
 reduces the deviation caused by faults, lowering the probability of the fault leading to SDCs. 
2) The ACT function is frequently used in different layers in DNNs (e.g., VGGNet, ResNet, etc), and thus applying range restriction on ACT function effectively dampens fault amplification in between layers (as DNNs usually have many layers). 

While range restriction in ACT function would limit the fault amplification effect, it is not sufficient as there are still computations that occur between the ACT functions. A single fault in these can quickly propagate and be amplified.  
For example, consider the following operation in a DNN.

\begin{equation}
\begin{split}
\label{eq:relu}
y = \text{Relu}_2( \text{Conv}(\text{MaxPool}(\text{ReLu}_1(x)), I)),
\end{split}
\end{equation}
 
where $\text{ReLu}_1$ and $\text{ReLu}_2$ are guarded by \sysname. Let their bounds be $bound(\text{ReLu}_1)=10, bound(\text{ReLu}_2)=1000$. For simplicity, assume $x=[1,2]$, and $\text{MaxPool}(\text{ReLu}_1(x))=2$. Let the Conv layer have $n$ kernels (we assume that each kernel is a simple 1x1 identity kernel $I$), each of which performs 
a dot-product computation. Thus the dimension of $y$ is $(n,1)$. 

Assume that a fault occurs at MaxPool function and deviates the MaxPool's output from $2$ to $1024+2$. The faulty value of $1026$ subsequently propagates through the Conv layer, and thus \emph{all} the elements from $y$ would be affected. In this case, the fault-free output of $y$ is a n-element vector of $2$. \sysname can restrict the values from 1026 to 1000 - thus $y$ becomes a vector of $1000$. Such a large value deviation has a higher probability of resulting in SDCs. Thus, applying \sysname on the ACT layer alone is \emph{not} enough to mitigate critical faults.

By analyzing the value dependency between layers in the model, we find the restriction bound applied to the ACT function can also be \emph{extended} to other functions. Using the same example above, the bound of ReLu$_1$ is also applicable to the MaxPool function, i.e., $bound(\text{MaxPool}(\text{ReLu}_1(x))) = bound(\text{Relu}_1(x))$. Therefore, the values from MaxPool should not be greater than $bound(\text{ReLu}_1)=10$, and thus $y$ will be a vector of $10$ even under the the fault in the MaxPool function, which has a significantly lower deviation (compared with a faulty vector with values of $1000$). As a result, the fault is less likely to cause an SDC as the limited deviation can be tolerated by the inherent resilience of DNNs. Therefore, we need to apply range restriction to selected layers beyond just the ACT function to effectively mitigate SDCs.

We describe the procedure of applying \sysname to an unprotected DNN in Algorithm~\ref{alg:ranger}. The input to the algorithm is the restriction bounds collected from the profiling process ($j$ pair of upper and lower bounds in total for $j$ ACT layers). The resulting output is the DNN protected with \sysname. 
Line 2 traverses each operation in the network. For each of the ACT operations, the output will be bounded (Line 3-4). 
For all the operations that follow (connect to) the ACT operation 
and belong to \{Max-Pool, Avg-Pool, Reshape, Concatenate\},  the \emph{same} restriction bounds  will be applied to their outputs as well (Line 5-8). These are the operations where \sysname can be deployed beyond the ACT operation (i.e., the operators protected by \sysname remain in the same network).

Note that for the Concatenate operation that concatenates the output from the previous 2 ACT operation (this is used in the SqueezeNet model), the restriction bound is derived from the bounds in the preceding 2 ACT operation: lower bound = $\text{min}(low_{j-1}, low_j)$, and upper bound = $\text{max}(up_{j-1}, up_j)$. 
The time complexity of the algorithm is $O(n)$, where $n$ is the network size (in nodes).

\begin{algorithm}[t]
\caption{Range restriction on DNNs}
\label{alg:ranger}
\begin{small}

\hspace*{\algorithmicindent} \textbf{Input:} $i \leftarrow$ number of operation in the model \\ 
\hspace*{1.35cm} $j \leftarrow$ number of ACT operation \\
\hspace*{1.35cm} $(low_j,up_j) \leftarrow$ bounds for the $j_{th}$ ACT op \\

\hspace*{\algorithmicindent} \textbf{Output:} Protected DNN with \sysname

\begin{algorithmic}[1]
\STATE /* Traverse each operation in the network from the first layer to the last one */
\FOR{$op_i$  \textbf{in} operations in the network} 
 
\IF{$op_i$ is the $j_{th}$ ACT operation} 
\STATE Bound $op_i$ with $(low_j, up_j)$ 
\IF{$op_{i+1}$ \textbf{in} \{Max-Pool, Avg-Pool, Reshape\} }
\STATE Bound $op_{i+1}$ with $(low_j, up_j)$ 
\ELSIF {$op_{i+1}$ \textbf{in} \{Concatenate\} }
\STATE Bound $op_{i+1}$ with $(\text{min}(low_{j-1}, low_j), \text{max}(up_{j-1}, up_{j}))$ 
\ENDIF
\ENDIF

\ENDFOR
\STATE \textbf{return} Protected DNN

\end{algorithmic}
\end{small}
\end{algorithm}


%% file: sections/implementation.tex
\label{sec:implementation}

We have implemented \sysname using the TensorFlow framework~\cite{abadi2016tensorflow}, which is the most popular framework for writing ML programs today~\cite{tf-popularity}. 
This allows \sysname to be applied to a diverse set of ML programs. 
In TensorFlow, the program is abstracted as a data-flow graph, which is composed of a set of operators (i.e., different functions). 
\sysname modifies the TensorFlow graph by adding the extra operators for range restriction (tf.minimum and tf.maximum) 
 as per Algorithm~\ref{alg:ranger}. 
 
 {We have publicly released the Ranger code under an open-source license (MIT) at the following URL:  \emph{https://github.com/DependableSystemsLab/Ranger}}

\added{We provide a Python script to \emph{automatically} insert \sysname into the DNNs (written as TensorFlow programs). 
The only inputs needed from the programmer are  (1) the restriction bound values, and (2) the name of the activation function such as Relu. The former can be obtained via profiling (Section~\ref{sec:experimental-setup}), while the latter is based on the specific DNN model. Note that it is often trivial to determine the activation function used in a DNN as it is a fundamental design parameter.} 
 
\sysname duplicates the TensorFlow graph, and inserts the range checks at the appropriate places, which is then imported into the existing graph. The reason it duplicates the graph is because TensorFlow has a static graph structure and hence the existing operators are not mutable (i.e., the graph is \emph{append-only}). We use TensorFlow's \emph{import\_graph\_def} functionality for this purpose. We insert the bounded operators via the \emph{input\_map} parameter that controls the input mapping relation. 

\begin{figure}[t]
\centering
  \includegraphics[height=1.9in]{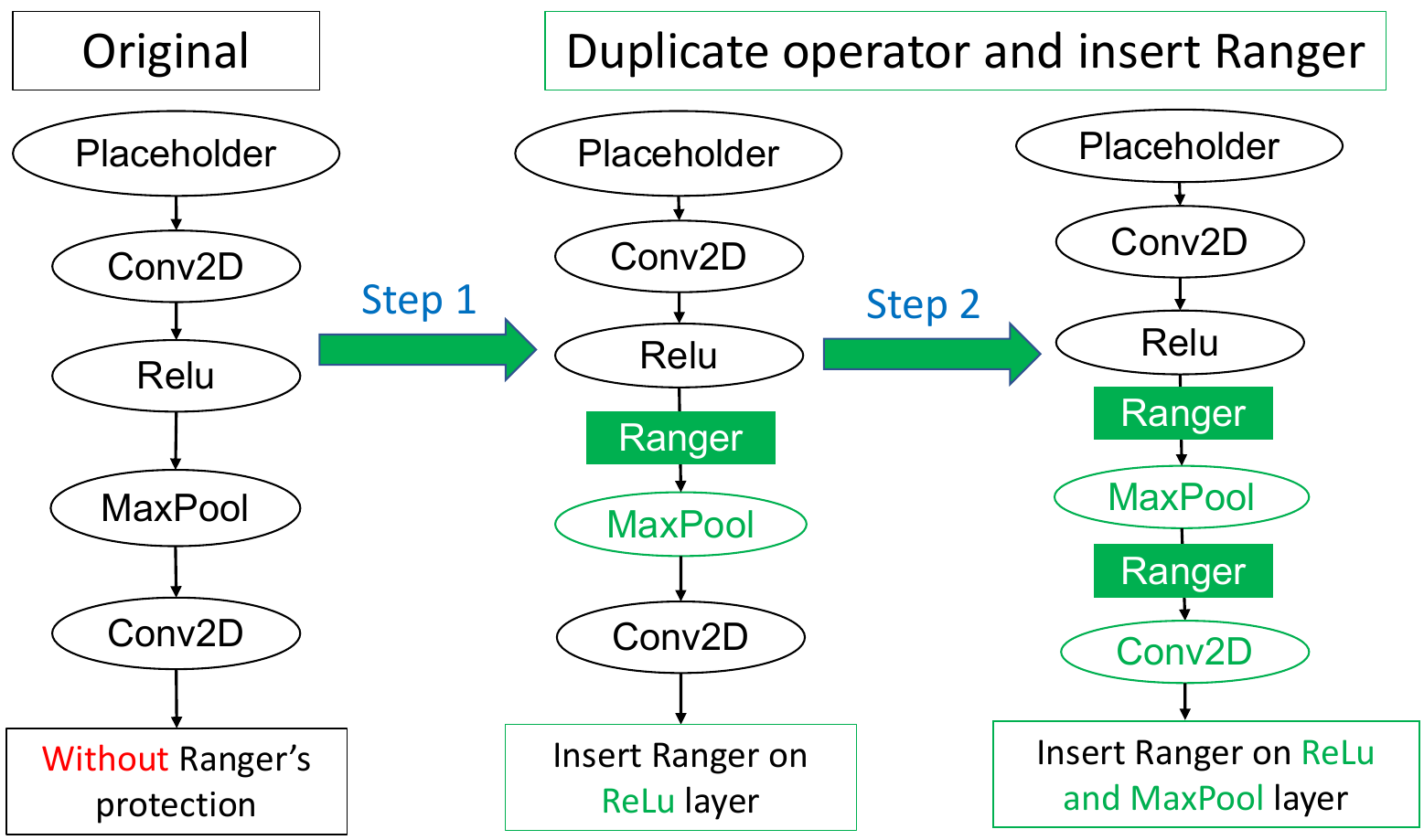}
\caption{Illustration of how to automatically deploy \sysname on a DNN by duplicating operators (model represented as a data-flow graph). }
\label{fig:ranger-implt}
\end{figure}

Fig.~\ref{fig:ranger-implt} shows an example. As shown in the left side of Fig.~\ref{fig:ranger-implt}, \emph{Relu} is connected as input to the next operator, \emph{MaxPool}. During graph duplication, \added{the output from \emph{Relu} is restricted with \sysname, which creates a new operator to be fed as input to the duplicated  MaxPool (outlined in green).}  
Each iteration of the duplication process applies \sysname in different DNN layers (Algorithm~\ref{alg:ranger}). In the example, Step-1 applies \sysname on the \emph{Relu} layer and Step-2 applies it on the \emph{MaxPool} layer.

Note that \sysname can also be applied to DNNs written in other ML frameworks such as PyTorch.
This is because \sysname leverages the inherent properties of the DNN model (e.g., the monotone property of the DNN components, the inherent resilience of DNN), which are platform-independent.  

%% file: sections/evaluation.tex
\label{sec:experiment}

\subsection{Experimental Setup}
\label{sec:experimental-setup}

\textbf{Fault Injection (FI) tool.}
Because \sysname works on TensorFlow, we use TensorFI \footnotemark[3], an open-source FI tool for TensorFlow-based programs that allows faults to be injected into the TensorFlow graph~\cite{li2018tensorfi}.  
{We inject random single bit-flip faults directly into the output values of operators in the graph, and observe the final output}. This is in line with the FI method in most prior work in the area \cite{schorn2018efficient,li2017understanding,chen2019,li2018tensorfi}.

Note that we only consider those faults that would not lead to obvious system failures 
such as a crash (e.g., modifying the dimension of a tensor might cause an error, and terminate the program), as such faults do not result in SDCs. 

\textbf{Hardware}: Our experiments were conducted on the following machines: 1) an Ubuntu Linux 18.04.2 system with, 8 RTX 2080Ti GPUs, 24 CPUs with 256 GB memory; 2) a Fedora Linux 20 system, 2 GTX TITAN GPUs, 16 CPUs with 256 GB memory; 3) an Ubuntu Linux 16.04 system with 6 CPUs, 1 GeForce GT610 GPU with 16 GB memory.

\textbf{Research Questions}: We evaluate \sysname by asking four research questions (RQs):\\
\textbf{RQ1:} What is the effectiveness of \sysname in fault correction in terms of reducing the SDC rates?  \\
\textbf{RQ2:} Does \sysname affect the accuracy of the model? \\
\textbf{RQ3:} What is the performance and memory overhead associated with \sysname? \\
\textbf{RQ4:} Is \sysname effective in DNNs using reduced precision data types?

\textbf{DNN benchmarks and datasets.}
In our evaluation, we evaluate DNNs applications in both classification and regression tasks. We consider common DNN classification benchmarks such as VGGNet, SqueezeNet, ResNet.
For the DNNs in regression tasks (where the output is a variable value instead of a class label), we choose two DNN applications that can be used in the AV domain - these are the applications that can predict the steering angles of the AV. We choose the Nvidia Dave driving model~\cite{bojarski2016end} and the steering model from Comma.ai~\cite{commaai} (which is an AV company). 
 These DNN applications have been adopted in real-world vehicles~\cite{dave-test}, and are  often used as benchmarks for AV studies~\cite{pei2017deepxplore,chen2019}.

We use standard ML datasets such as MNist, Cifar-10, ImageNet), as well as a real-world driving dataset collected from the driving scenes captured by a real vehicle~\cite{drivingDataset}.
%


Table~\ref{tab:eval} summarizes the models and datasets in our evaluation. For models using the ImageNet dataset, we report both the top-1 accuracy (i.e., the target label is the predicted class that has the highest probability) and the top-5 accuracy (i.e., the target label is one of the top 5 predictions)~\cite{li2017understanding}. 
For the two steering models, we use RMSE (root mean square error) and the average deviation per frame to evaluate the model's accuracy - these are also used in other AV DNN studies~\cite{du2017self}.


\begin{table}   
\caption{DNN models and datasets used for evaluation}
\label{tab:eval}
\centering 
\footnotesize
\renewcommand{\arraystretch}{1}
\begin{tabular}{|c|c|c|}
\hline
{\bf DNN model} & {\bf Dataset} & {\bf Dataset Description}\\
\hline
LeNet & MNist & Hand-written digits \\
\hline
AlexNet & Cifar-10 & General images  \\
\hline 
VGG11 & GTSRB & Real-world traffic sign \\
\hline
VGG16 & ImageNet & General images  \\
\hline 
ResNet-18 & ImageNet & General images \\
\hline 
SqueezeNet  & ImageNet & General images \\
\hline 
Nvidia Dave~\cite{autopilot-git} & Driving & Real-world driving frames \\
\hline 
Comma.ai~\cite{commaai} & Driving & Real-world driving frames \\
\hline
\end{tabular}
\end{table} 

\footnotetext[3]{https://github.com/DependableSystemsLab/TensorFI}


Note that FI experiments on DNNs are highly time-consuming - for each input, we need to perform thousands of FI experiments to obtain a statistically significant estimate of the SDC rate. 
To balance the experimental time with coverage, we choose 10 inputs per model, and ensure that the DNNs are able to generate correct predictions on these inputs (in the absence of faults). We perform $5000$ FI trials for each DNN, except for the DNNs using ImageNet, where we perform only $3000$ FI trials as they take longer to run (we verified that the results are statistically significant). We also calculated the standard error bars at the 95\% confidence level. 

In our evaluations, we use 32-bit fixed-point data type for the first 3 RQs (this is more energy-efficient than using a floating-point data type \cite{judd2018proteus}). 
For RQ4, we evaluate \sysname on DNNs using a 16-bit fixed-point data type. 

\begin{figure}[t]
	\centering
	\includegraphics[width=2.5in, height=1.8in]{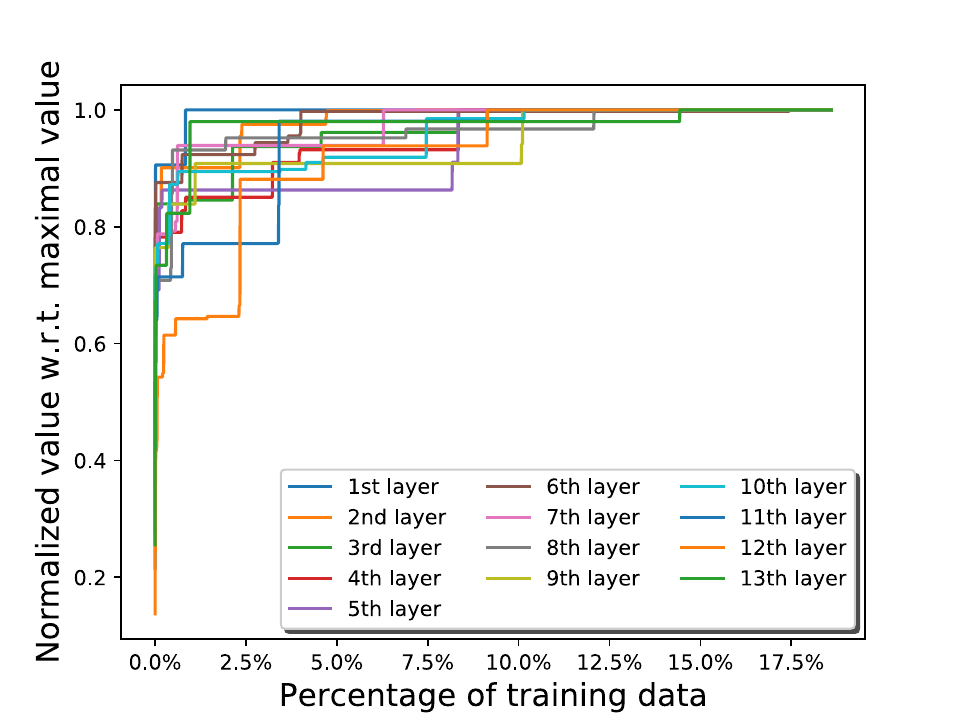}
	\caption{Ranges of values observed in each ACT layer using different amount of data on the VGG16 network (13 ACT layers in total). A total of 186056 images (around 20\% of the training set) were used.}
	\label{fig:cdf-vgg}
\end{figure}

\textbf{ {Deriving} Restriction Bounds.}
In our work, we  {derive} the restriction bounds from a randomly-sampled subset of the \emph{training set}. 
We choose 20\% of the training data from each dataset. 
We find that this is sufficient to  {derive} the ranges for all the DNNs used in our study.   
For example, Fig.~\ref{fig:cdf-vgg} shows the ranges of activation values obtained when sampling different amounts of training data on the VGG16 model. The values are normalized with respect to the global maximal values (i.e., maximal values on all the sampling data). As shown, the value range quickly converges to the global maximal values for all layers. We observe similar trends in the other DNNs, but do not report them due to space constraints. 

Note that  {deriving} restriction bounds is a one-time cost, and is \emph{incurred before the deployment of the DNN}.  {The exact time is dependent on the size of the training data and the network. }
In our experiments, it took around one hour to complete this process on the largest network (VGG16) in our evaluation with around 20\% of training data in the ImageNet dataset. 


To reduce \emph{Ranger's} effect on the model's accuracy, we conservatively choose the maximal value observed during the sampling process as the restriction bound (we study the effect of choosing different restriction bounds in Section~\ref{sec:discussion}). 

\begin{figure}[t]
\centering
  \includegraphics[width=3.4in, height=0.9in]{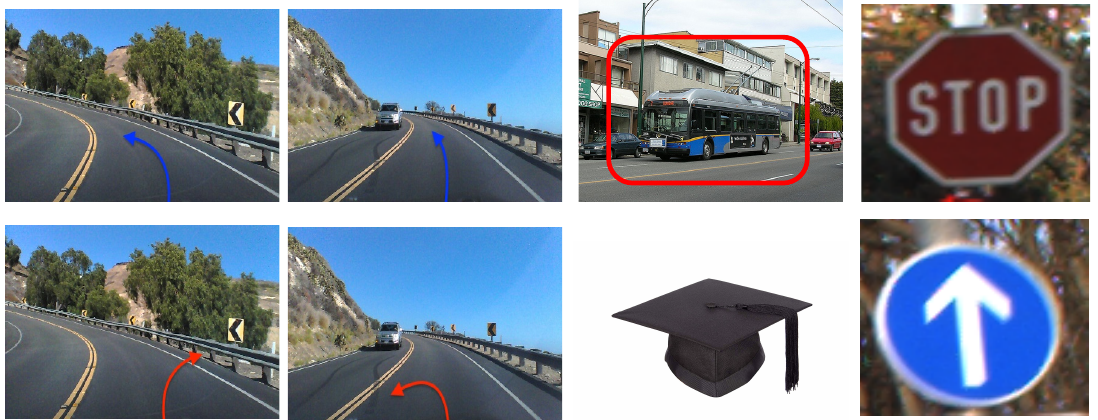}
\caption{Examples of SDCs in different DNNs in our evaluation. The first row shows the input to the DNN, the second row shows the corresponding output produced by the DNN under faults.}
\label{fig:sdc-effect}
\end{figure}

\subsection{Results}
\label{sec:results}
\textbf{RQ1: Effectiveness of range restriction.} We measure the \emph{SDC rate}, which is the percentage of the transient faults that cause SDCs, with and without \sysname. For classifier models, an SDC will manifest itself as an image misclassification. For the two steering models that produce continuous values as outputs, we use different threshold values for the deviations of steering angles to identify SDCs, namely 15, 30, 60 and 120 degrees as done by our prior work~\cite{chen2019}. Fig.~\ref{fig:sdc-effect} illustrates the effect of SDCs in different DNNs tasks from our experiments. 

{Though \sysname selectively applies range restriction on a subset of operations in the DNN, we consider faults that can arise randomly in \emph{all} the operations during the computations except the last FC layer.}
We exclude the last FC layer because values in this layer are directly associated with the final outputs. Thus, restricting the values in the last FC layer is not effective in mitigating SDCs (we validated this in our experiments). However,  the state space of the last FC layer constitutes a very small fraction of the state space (e.g., in VGG16 model, the last FC layer only accounts for 0.0047\% of the state space), and techniques such as duplication can be used to protect this particular layer with minimal overheads.

\begin{figure}[t]
\centering
  \includegraphics[width = 3.5in, height=1.1in, trim=4 4 4 4,clip]{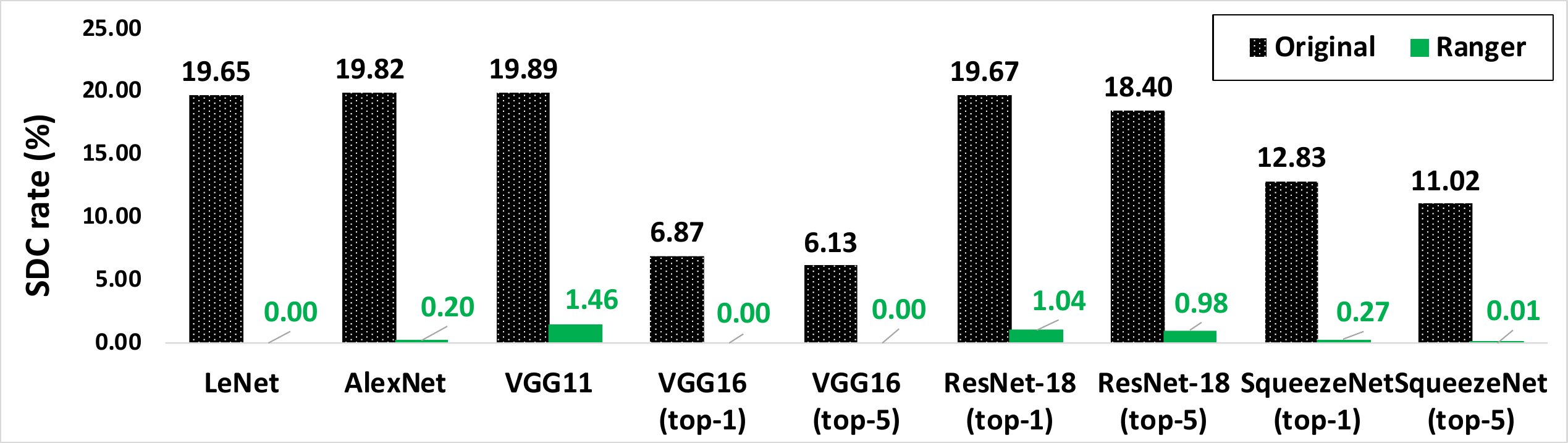}
\caption{SDC rates of the original classifier models, and models protected with \sysname. For the models using ImageNet, we show the results for top-1 and top-5 accuracy. Error bars range from $\pm0.04\%$ to $\pm1.46\%$ at the 95\% confidence interval. Lower values are better.}
\label{fig:sdc-1}
\end{figure}

\begin{figure}[t]
\centering
  \includegraphics[width = 3.5in, height=1.1in, trim=4 4 4 4,clip]{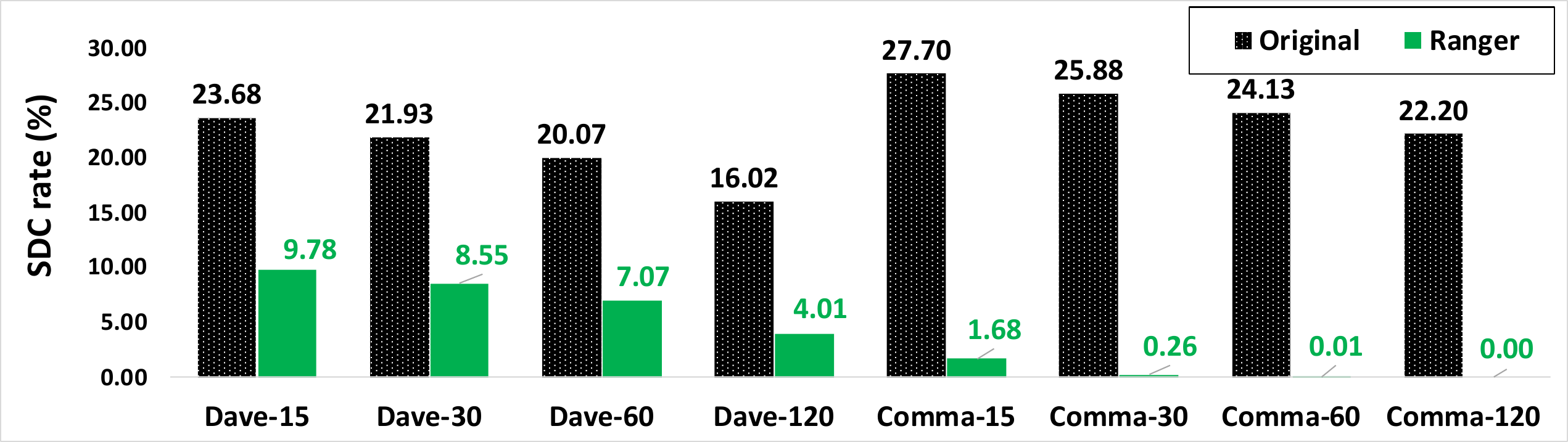}
\caption{SDC rates of the original steering models, and models protected with \sysname. An SDC is determined based on different thresholds of the degrees of deviation to the correct steering angles (15, 30, 60, and 120 degrees). Error bars range from $\pm0.03\%$ to $\pm1.24\%$ at the 95\% confidence interval. Lower values are better.}
\label{fig:sdc-2}
\end{figure}

Fig.~\ref{fig:sdc-1} illustrates the SDC rates in the 6 classifier models with and without \sysname. While different DNNs exhibit different SDC rates, Fig.~\ref{fig:sdc-1} shows that \sysname achieves significant SDC reduction across all the models.
For example, in the LeNet model, the SDC rate decreases from around $20\%$ to $0\%$ (in our experiments). {\em On average, \sysname reduces the SDC rates from $14.92\%$ to $0.44\%$ across the models.}

The results from the two steering models are shown in Fig.~\ref{fig:sdc-2}. In the Comma steering model, \sysname can reduce almost all the deviations of steering angles due to transient faults, and completely eliminate large deviations (e.g., $0\%$ of SDCs in the category of threshold=120). However, \sysname achieves less pronounced SDC reduction in the Dave model. This is because the Dave model outputs the steering angle in radians, while the Comma model outputs the steering angle in degrees.  The conversion from degrees to radians is more sensitive to deviations. The reason is that the conversion function (Atan function in TensorFlow) is horizontal asymptote (y$\in$($-\pi/2, \pi/2$)), and thus even a small deviation at the input of Atan function would cause a large output deviation, i.e., higher SDC probability.
Based on this observation, we train a new model which outputs the steering angle in degrees instead of radians, which achieves both {\em better accuracy and better resilience} with \sysname than the original model (Section~\ref{sec:discussion}).

\emph{Quantitative Comparison with related work.} Hong \emph{et al.}~\cite{hong2019terminal} suggest a defense mechanism against memory bit-flips by modifying the ACT functions of the models such as changing ReLu into Tanh (a similar approach is also proposed in \cite{hoang2019ft}). 
We compare \sysname with the method in Hong \emph{et al.} for 5 of the 8 DNNs. The models we consider are: LeNet, AlexNet, VGG11, Dave and Comma steering model.
We only consider these 5 models as we need to train a new model for each DNN, and it is time consuming to do so for the other 3 DNNs. 

\begin{figure}[t]
\centering
  \includegraphics[width = 3.5in, height=1.3in, trim=2 2 2 2,clip]{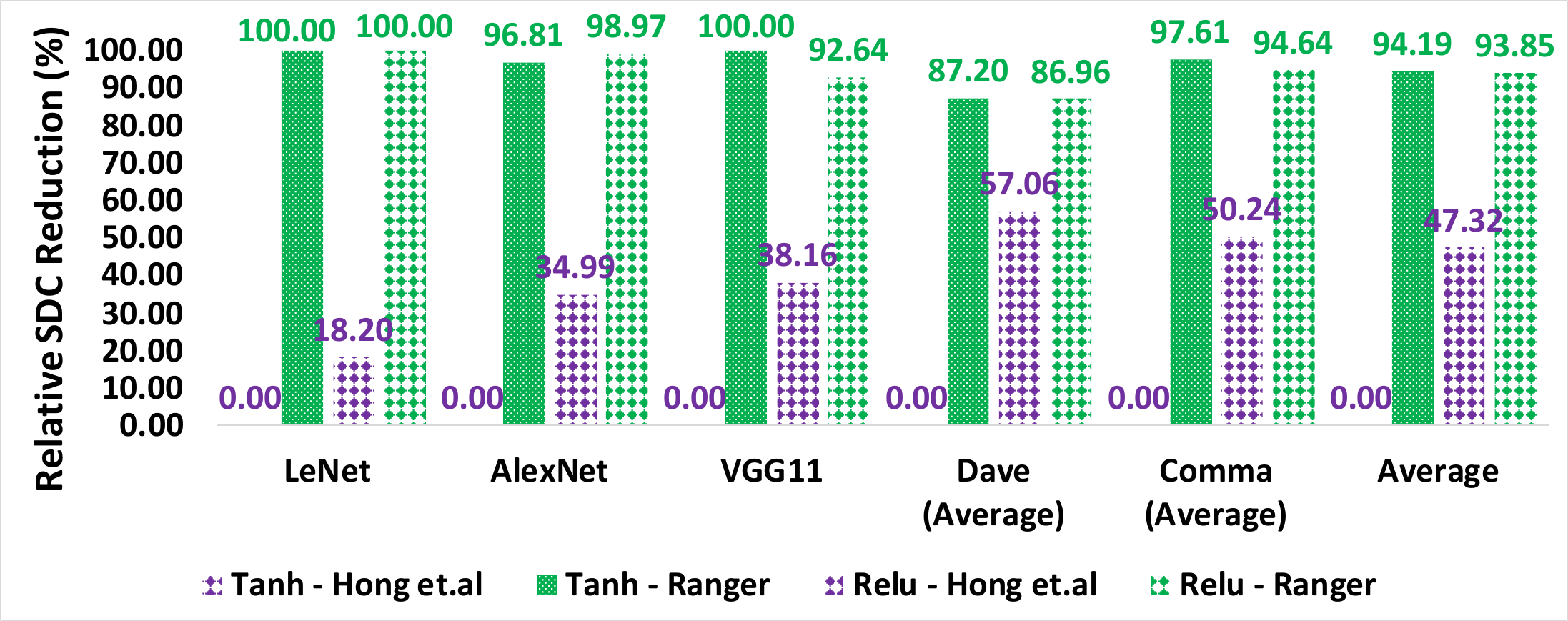}
\caption{SDC rate reduction by protecting the DNN with the approach in Hong \emph{et al.} \cite{hong2019terminal} and \sysname. Error bars range from $\pm0.12\%$ to $\pm1.38\%$ at the 95\% confidence interval. Higher values are better.}
\label{fig:marland-comp}
\end{figure}

For both approaches, we report the SDC rate reduction relative to the original SDC rates in Fig.~\ref{fig:marland-comp} (for brevity, we report the average results for the steering models across all the thresholds). 
Fig.~\ref{fig:marland-comp} shows that the approach from Hong \emph{et al.} achieves 0\% relative SDC reduction in models using the Tanh activation function. This is because transient faults could occur after the Tanh function in the network, and these would not be affected by the replacement.
%
In contrast, \sysname achieves significant SDC reduction because it performs selective range restriction in the entire DNN. 

For models using the ReLu function, while both approaches can reduce SDC rates, \sysname enables significantly higher resilience boosting than Hong \emph{et al.}. For example, when $threshold=120$ in the two steering models, the SDC rates in Hong \emph{et al.} vary from $4.76\%$ to $9.48\%$, while with \sysname, the SDC rates vary from $0\%$ to $0.27\%$, which is {\em an order of magnitude} lower. 
 On average, \sysname achieves \emph{more than 90\% SDC rate reduction than Hong \emph{et al.}} across all the models. 


\begin{framed}
 \sysname reduces the SDC rate from $14.92\%$ to $0.44\%$ (34X reduction) for the 6 classifier DNNs, from $24.98\%$ to $0.49\%$ on the Comma.ai model (50X reduction), and $20.42\%$ to $7.35\%$ (2.77X reduction) on the Dave model. 
\end{framed}

\textbf{RQ2: Accuracy.} { 
In this RQ, we investigate whether the value truncation by \sysname affects the accuracy of the DNNs in the absence of faults. 
We follow the common practice in ML studies to \emph{separate the training and validation set}. The latter is a set of data 
used for simulating the unseen data, and for evaluating the model learned from the training set. 
Because \sysname derives the restriction bounds from a subset of the \emph{training set}, we choose to evaluate the models on the \emph{validation set}. 
We compare the accuracy of the original models both with and without \sysname on the validation set, in the absence of faults.
The results are presented in Table~\ref{tab:acy-comp}.} 

Our results show that applying \sysname does not degrade the accuracy of the baseline models in any of the 8 DNNs. This is because the restriction bounds derived from existing data are sufficient to characterize the ranges of values in the DNN (Figure~\ref{fig:cdf-vgg}). For the rare cases where the normal values exceed the restriction bounds, the value reduction due to \sysname does not affect the accuracy. 

In the SqueezeNet model, applying range restriction marginally increases the accuracy of the model. This is because the large value corresponding to the incorrect label is reduced by \sysname, and thus the classification probability of the incorrect label also decreases. 


\begin{table}[t]
\caption{Accuracy of original DNN, and the DNN protected with \sysname. $+$ indicates accuracy improvement. Higher values are better.}
\label{tab:acy-comp} 
\begin{center}
\footnotesize
\renewcommand{\arraystretch}{1}
\begin{tabular}{|c|c|c|c|}
\hline
{\bf DNN model} & {\bf w/o \sysname} & {\bf w/ \sysname} & {\bf Diff.}\\
\hline
LeNet & 99.20\% & 99.20\% & 0.00\% \\
\hline
AlexNet & 82.14\% & 82.14\% & 0.00\%\\
\hline 
VGG11 & 99.74\% & 99.74\% & 0.00\% \\
\hline 
VGG16 (top-1) & 64.72\%  & 64.72\%  & 0.00\%\\
VGG16 (top-5) & 85.736\%  & 85.736\% & 0.00\%\\
\hline 
ResNet-18 (top-1) & 62.66\% & 62.66\% & 0.00\%\\
ResNet-18 (top-5)	& 84.61\%  & 84.61\% & 0.00\%\\
\hline 
SqueezeNet (top-1) & 52.936\%  & 52.940\%  & +0.004\%\\
SqueezeNet (top-5)	& 74.150\%  & 74.154\%  & +0.004\%\\
\hline
Dave (RMSE) & 9.808 & 9.808  & 0.000\\
Dave (Avg. Dev.)	& 3.153  & 3.153 & 0.000 \\
\hline 
Comma (RMSE) & 24.122 & 24.122 & 0.000\\
Comma (Avg. Dev.)	& 12.640  & 12.640  & 0.000\\
\hline
\end{tabular}
\end{center}
\end{table} 

\begin{framed}
\sysname does not degrade the accuracy of any of the evaluated DNN models in our experiments.
\end{framed}

\textbf{RQ3: Overhead.} 
 {
There are two sources of overhead due to \sysname: (1) instrumentation overhead, and (2) runtime overhead. The instrumentation overhead is an one-time effort incurred before the model deployment, while the runtime overhead is incurred during the deployment of the model.}
	
 {We first measure instrumentation overhead to automatically insert \sysname into the models.  The instrumentation time for each model is shown in Table~\ref{tab:instrument-time} (on a MacBook Pro laptop with 2.3 GHz Intel Core i5 and 16 GB memory). Overall, \sysname takes an average of 46 seconds for instrumentation. 
}
\begin{table}[t]
\caption{Time taken to automatically insert \sysname in the models.}
\label{tab:instrument-time}
\begin{center}
\footnotesize
\renewcommand{\arraystretch}{1}
\begin{tabular}{|c|c|c|c|}
\hline
{\bf DNN} & {\bf Insertion time} & {\bf DNN} & {\bf Insertion time}\\
\hline
LeNet & 3 sec & ResNet-18 & 22 sec \\
\hline
AlexNet & 2 sec & SqueezeNet & 11 sec\\
\hline 
VGG11 & 7 sec & Dave & 2 sec \\
\hline 
VGG16  & 320 sec  & Comma  & 1 sec \\
\hline 

\end{tabular}
\end{center}
\end{table} 

We measure the runtime overhead of \sysname in terms of its memory and performance overheads during the inference phase. The memory overhead comes from the storage for the restriction bounds, which is proportional to the number of ACT functions in the models (the restriction bounds are applicable for not just ACT functions, but also others such as MaxPool function). Given the typical size of DNNs, this overhead is negligible (e.g., VGG16 has a size of over 500MB).  

{For the runtime performance overhead, we first measure the absolute runtime of the DNN models. The average inference time across all the models (using GPUs) is 9.41 milliseconds (without \sysname) and 9.64 milliseconds (with \sysname), with $\pm0.11\%$ standard deviation. Note that the above runtime is highly dependent on the hardware architecture and system configuration. 
}
Therefore, for reproducibility purposes, we measure the performance overhead of \sysname in terms of the floating-point operations (FLOPs) incurred by it (using TensorFlow's profiler). 
FLOPs is a measure of the latency and energy consumption of ML models~\cite{tang2018experimental, han2015learning, sehgal2019guidelines},
 and is independent of the platform. 

\begin{table}[t]
\caption{Computation overhead of \sysname measured in FLOPS. M stands for Million; B stands for Billion.}
\label{tab:runtime}
\begin{center}
\footnotesize
\renewcommand{\arraystretch}{1}
\begin{tabular}{|c|c|c|c|}
\hline
{\bf DNN model} & {\bf w/o \sysname} & {\bf w/ \sysname} & {\bf Overhead}\\
\hline
LeNet & 24.622M & 24.724M & 0.412\% \\
\hline
AlexNet & 11.361M & 11.484M & 1.082\%\\
\hline 
VGG11 & 87.057M & 87.326M & 0.309\% \\
\hline 
VGG16  & 309.604B  & 309.905B  & 0.097\%\\
\hline 
ResNet-18  & 36.354B & 36.404B & 0.138\%\\
\hline 
SqueezeNet  & 530.813M  & 539.215M  & 1.583\%\\
\hline
Dave  & 56.545M & 56.764M & 0.387\% \\
\hline 
Comma  & 17.673M & 17.714M & 0.235\%\\
\hline

\end{tabular}
\end{center}
\end{table}


The results are reported in Table~\ref{tab:runtime}.  
We did not observe variations across different inputs as they all have the same dimension. 
We find that the overhead of \sysname to be very low (0.530\% on average) - this is negligible in most cases. 
This is because \sysname only involves range checking and truncation, which are relatively simple operations, compared to the typical operators in a DNN such as convolution.

\begin{framed}
\sysname incurs negligible performance and memory overheads on average across the DNN models. 
\end{framed}

\textbf{RQ4: Effectiveness of \sysname under reduced precision data type.} In previous RQs, we used a 32-bit fixed-point data type, which is more energy-efficient than a floating-point data type. The data type precision can be reduced to gain further energy efficiency~\cite{judd2018proteus}. Therefore, we study the effectiveness of \sysname in DNNs using 16-bit fixed-point data type, which can be used for inference in large DNNs. Using smaller data types such as 8-bit datatype cannot provide sufficient dynamic value range for large DNNs~\cite{judd2018proteus},  
{
	and hence we do not study these. 
Nevertheless, data quantization with lower bit widths is similar to using standard fixed-point datatypes without quantization (e.g., a quantized 8-bit datatype represents comparable range as an unquantized 16-bit datatype).
Thus, we posit that \sysname will still be effective on them because the observation that the critical faults needs to yield large deviation is independent of the implementation.
}
We use 14 bits for the integer and 2 for the  fractional part, which is sufficient for common DNNs \cite{judd2018proteus}. 

\begin{figure}[t]
\centering
  \includegraphics[width = 3.4in, height=1.2in, trim=4 4 4 4,clip]{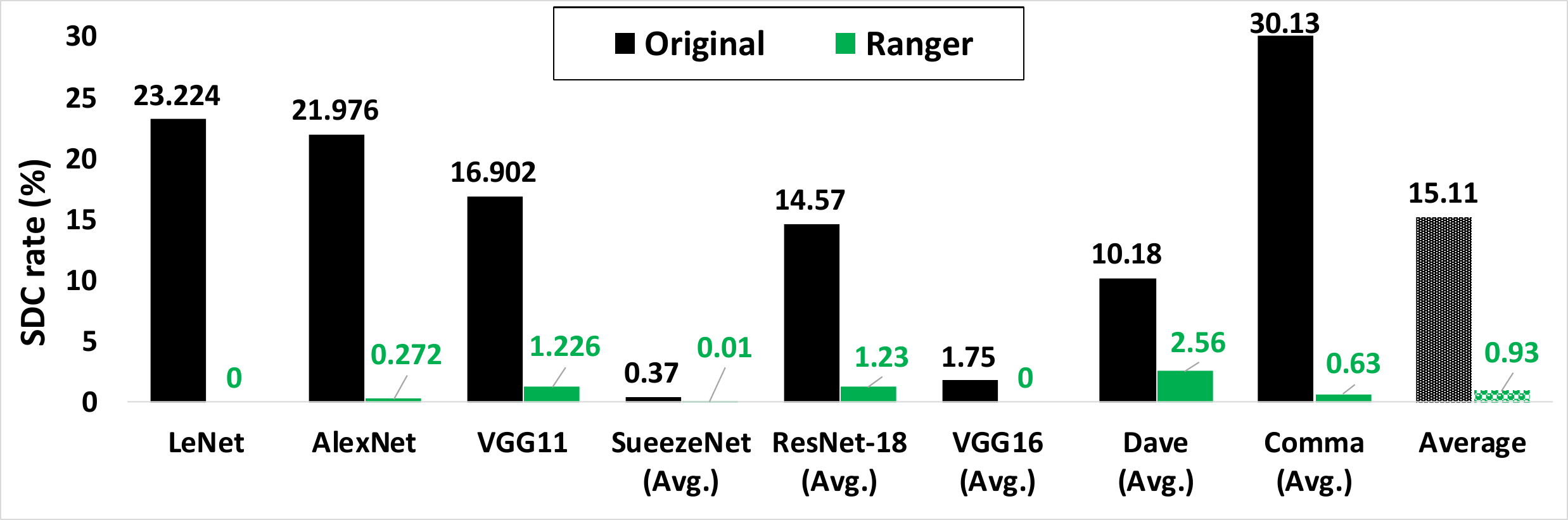}
\caption{SDC rate of DNNs using a 16-bit fixed-point data type, with and without \sysname. Error bars range from $\pm0.04\%$ to $\pm1.33\%$ at the 95\% confidence interval. Lower values are better.}
\label{fig:16-bit-res}
\end{figure}

As before, we report the SDC rates of the models with and without \sysname in Fig.~\ref{fig:16-bit-res}. As shown, \sysname is effective in reducing the SDC rate of all 8 DNNs, from $15.11\%$ to $0.93\%$ (a 16X reduction) even with reduced precision data types.

%% file: sections/discussion.tex
\label{sec:discussion}

In this section, we study the trade off between accuracy and resilience in \sysname, then the effectiveness of \sysname under multi-bit flips, and finally the design alternatives for \sysname. 

\subsection{Trade-off between Accuracy and Resilience.} 
\label{sec:trade-off-dave}
We study how to adjust the restriction bound to gain additional resilience at the cost of accuracy (e.g., in systems that are more prone to transient faults). We focus on the Dave steering model because as shown in Fig.~\ref{fig:sdc-2}, the average SDC rate is around $7\%$ even when it is protected with \sysname. 

As mentioned, \sysname does not yield significant SDC reduction on the original Dave model that outputs the radian value. Therefore, we train a new Dave model whose output is the steering angle in degrees. 
We then evaluate both the accuracy and SDC rate of the model using different restriction bound values. As mentioned in Section~\ref{sec:methodology}, we collect the distribution of ACT values from statistical sampling, and can choose the restriction bound accordingly. For example, setting the restriction bound to the $100^{th}$ percentile means we use the value that covers {\em all of the sampled values} (i.e., the maximum value) - this was our earlier approach in Section~\ref{sec:experiment}. 

\begin{figure}[t]
\centering
  \includegraphics[width = 3.5in, height=1.1in, trim=4 4 4 4,clip]{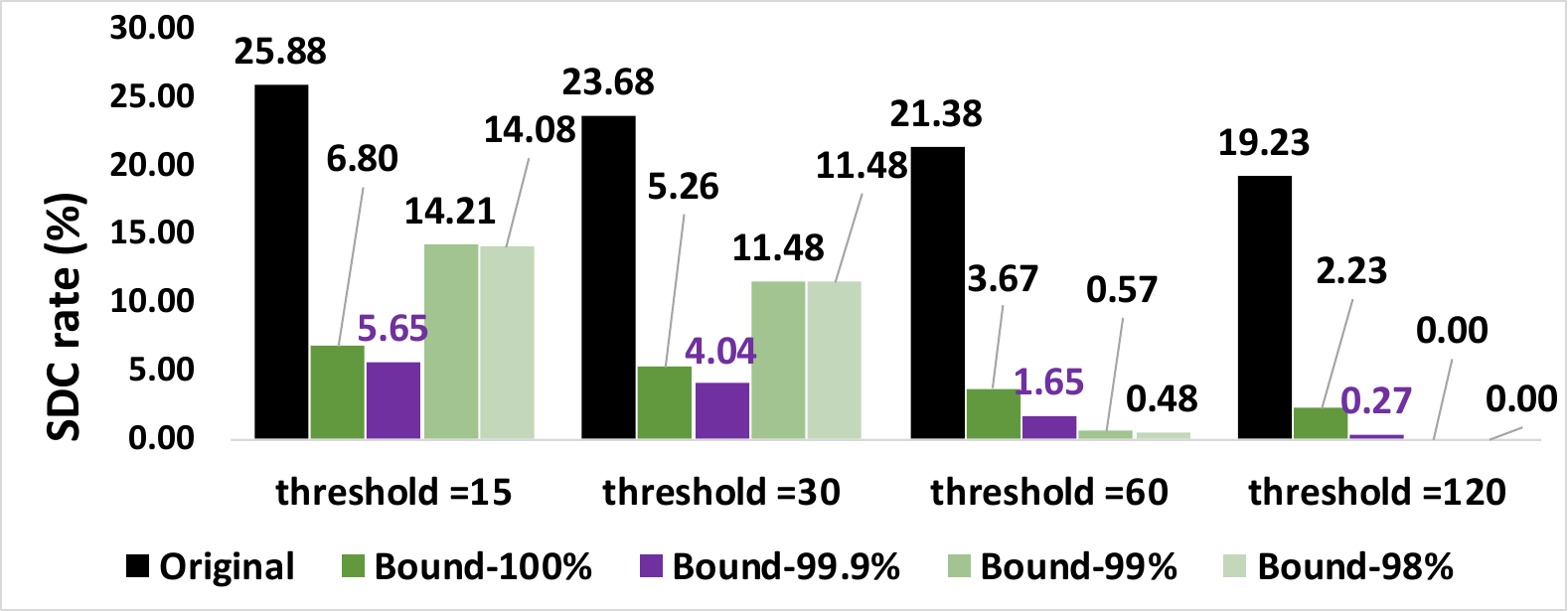}
\caption{SDC rates for the Dave model protected by \sysname for different restriction bounds. Error bars range from $\pm0.14\%$ to $\pm1.39\%$ at the 95\% confidence interval. Lower values are better.}
\label{fig:dave-tradeoff}
\end{figure}

Fig.~\ref{fig:dave-tradeoff} shows the SDC rates of the model with different restriction bound percentiles, and  Table~\ref{tab:dave-acy} shows the corresponding accuracy values.
As can be seen, the SDC rate reduction due to \sysname is higher than that in Fig.~\ref{fig:sdc-2}, leading to a lower SDC rate with \sysname. For example, when applying \sysname in both models, the SDC rate for the category of {\em threshold=120} is $4.01\%$ in the original Dave model, and $2.23\%$ in the new model. 
Also, \sysname does not degrade the accuracy of either model.
In fact, the accuracy of the new Dave model is higher than that in Fig.~\ref{fig:sdc-2}. This is because the new model outputs the steering angle in degrees, which has a larger dynamic range than radian values, 
thereby allowing the model to make more accurate predictions. 

As expected, setting a restriction bound to a lower-percentile value boosts the resilience, but also degrades the accuracy of the models. When comparing the models using $100\%$ bound and $99.9\%$ bound, the average deviation per frame increases from $2.651$ to $2.883$. 
However, this provides a higher resilience boost, e.g., the SDC rate for $threshold=120$ reduces from $2.23\%$ to  $0.27\%$. 
On average, choosing the 99.9\% percentile restriction bound reduces the SDC rate from $22.54\%$ to $2.9\%$ ($7.7x$ reduction), for this model, with marginal accuracy loss.

\begin{table}[t]
\small
\caption{Accuracy of the Dave model with different restriction bounds of \sysname. 100\% is the original. Lower values are better. }
\label{tab:dave-acy}
\begin{center}
\renewcommand{\arraystretch}{1}
\begin{tabular}{|c|c|c|c|c|c|}
\hline
{\bf Accuracy} & {\bf Original} & {\bf 100\%} & {\bf 99.9\%} & {\bf 99\%} & {\bf 98\%}\\
& &  Bound & Bound &  Bound & Bound \\
\hline
RMSE & 6.069 & 6.069 & 8.5719 & 12.370 & 13.940 \\
\hline
Avg. Deviation & 2.651 & 2.651 & 2.883 & 4.077 & 4.884 \\
\hline
\end{tabular}
\end{center}
\end{table}

\subsection{Fault Correction for Multi-bit Flips}
\label{sec:multi-bit-ranger}
We now evaluate the effectiveness of \sysname under the multi-bit flip fault model. 
{Note that multi-bit flips can manifest either as multiple, independent~\cite{sangchoolie2017one} bit-flips (i.e., in multiple values), or as consecutive bit-flips~\cite{yang2020practical} (i.e., in the same value). 
We assume the former model, namely multiple, independent bit-flips, as this will potentially result in \emph{more} values being affected by the fault. Therefore, our analysis is conservative. 
}

We hypothesize that \sysname will continue to be effective as it is agnostic to the number of out-of-bound values, as it restricts \emph{all} the values that exceed the restriction bound. 
{To test this hypothesis, we conducted experiments on two of the classifier models (LeNet and ResNet) and the two DNN models used in AVs. 
We inject multiple-bit flips (ranging from 2 bits to 5 bits) at randomly chosen values. These multi-bit faults can affect \emph{multiple} data values - thus, they are more potentially damaging as more values will be corrupted by faults.
As before, we inject 3000 faults per input per model.}

\begin{figure}[t]
\centering
  \includegraphics[width = 3.5in, height=1.1in, trim=2 2 2 2,clip]{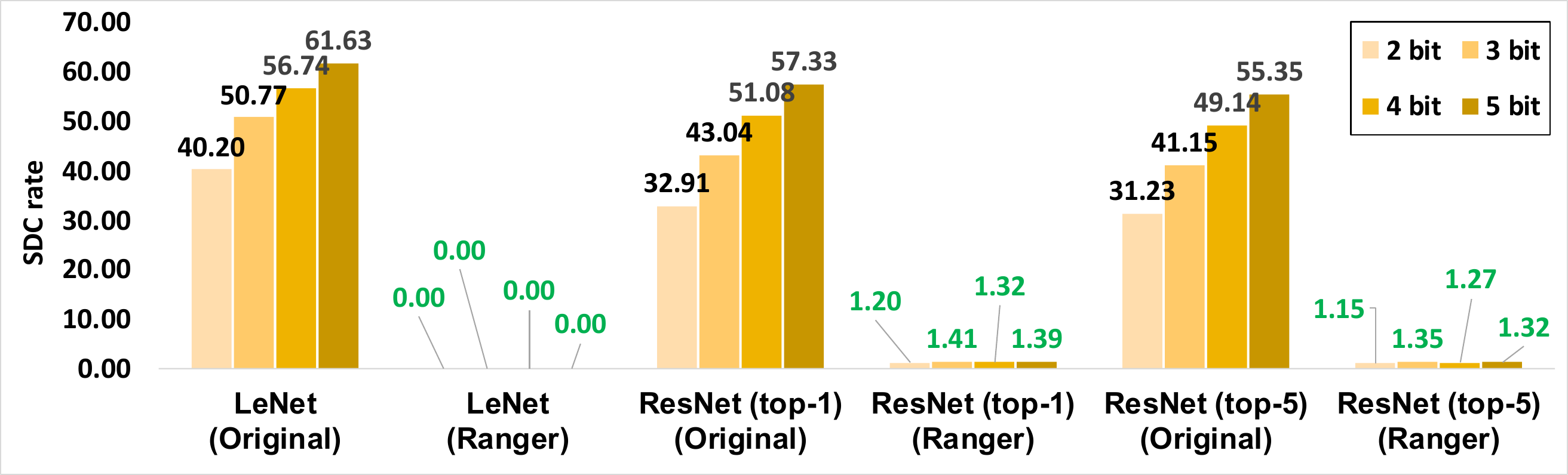}
\caption{SDC rates for the classifier models (with and without \sysname) under multi-bit flips. Numbers in green are the SDC rates of models protected by \sysname.
Error bars range from $\pm0.38\%$ to $\pm1.79\%$ at the 95\% confidence interval. Lower values are better.}
\label{fig:multi-bit}
\end{figure}

\begin{figure}[t]
\centering
  \includegraphics[width = 3.5in, height=1.1in, trim=2 2 2 2,clip]{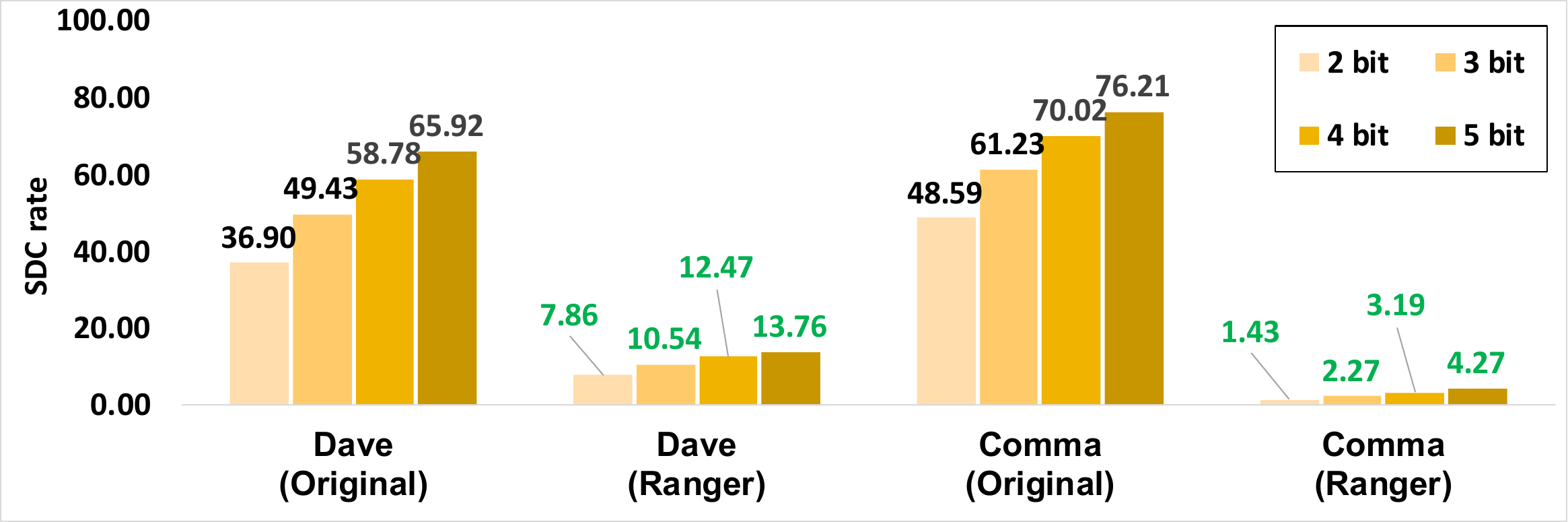}
\caption{SDC rates for the AV DNN models (with and without \sysname) under multi-bit flips. Numbers  in green are the SDC rates of the models protected by \sysname.
Error bars range from $\pm0.43\%$ to $\pm1.79\%$ at the 95\% confidence interval. Lower values are better.}
\label{fig:multi-bit-driving}
\end{figure}

{The results are shown in Fig.\ref{fig:multi-bit} and Fig.\ref{fig:multi-bit-driving}. As shown, the SDC rates of the original models increase with the number of multi-bit flips without \sysname. However, with \sysname, the SDC rates are significantly lower, in all cases. Further, the SDC rates for the classifier models protected by \sysname remain relatively constant with the number of bit-flips, while the SDC rates in the AV DNNs protected by \sysname increase with the number of bit flips (though at a slower rate than without \sysname). This is because the AV DNNs output the steering angle value, which require exactness (unlike classifier models). 
	
\emph{Overall, with multiple bit-flips, the average SDC rate is reduced from 47.55\% to 0.87\% (55X reduction) for classifier DNNs; and 58.38\% to 6.97\% (8.4X reduction) for AV DNNs.} 
}


\subsection{Design Alternatives for \sysname}
While \sysname restores out-of-bound values to the restriction bound, we explore other design alternatives for \sysname based on other work.
For example, Reagen \emph{et.al} propose to reset the faulty value to 0 upon the detection of a fault~\cite{reagen2016minerva}. 
Similarly, we can restrict all the out-of-bound value to 0, instead of the restriction bound. We conduct a targeted experiment on the VGG16 model using this strategy. 
We observe that resetting values beyond the bound to 0 significantly degrades the accuracy of the original model (e.g., $3/5=60\%$ of the inputs are inaccurate after the resetting). This is because the value reduction is so drastic that the model is not able to generate a correct output.  Further, resetting the values to 0 is likely to cause 0 values in subsequent operations in the network, such as multiplication, which can lead to incorrect results. 

Another possibility is to replace out-of-bound values with a random value between 0 and the restriction bound. We perform another experiment where we reset the values outside the restriction range to random values (within the restriction bound) - we inject $1000$ faults for this experiment. Even though some of the classification results change (where the incorrectly predicted labels change), the top-1 and top-5 accuracies remain the same. Therefore, random replacement  is also a viable strategy, though \sysname is much more deterministic and may be preferred for safety-critical systems such as AVs.

%% file: sections/relatedwork.tex
\label{sec:related-work}

\emph{Assessing the reliability of DNNs:}
Assessing the resilience of a system is often the first step towards protecting it. Therefore, there has been a plethora of work aiming to understand the reliability of DNNs via FI. Li \emph{et al.} assess the resilience of the DNNs by randomly injecting transient faults into the DNN accelerator~\cite{li2017understanding}. 
Reagen \emph{et al.} perform FI into the DNNs systems to study the trade off between model accuracy and fault rate~\cite{reagen2018ares}. Santos \emph{et al.} study the resilience of ML application under mixed-precision architectures~\cite{dos2019reliability}. In prior work, we 
propose a binary FI technique to effectively identify the critical faults in DNNs, based on the mathematical property of the DNNs' components~\cite{chen2019}, i.e., its monotonicity (similar to \sysname). 
While these studies provide insights into the ML applications' resilience, they do not   mitigate the faults.

\emph{Improving the reliability of DNNs:}
Several techniques have been proposed to enhance the error resilience of DNNs~\cite{li2017understanding,schorn2018efficient,hoang2019ft,zhao2020algorithm,ozen2019sanity}. 
Mahmoud \emph{et al.}~\cite{mahmoud2020hardnn}  use statistical FI to identify vulnerable regions in DNNs, and selectively duplicate the vulnerable computations for fault detection. 
However, their approach incurs high computational overheads. 
Li \emph{et al.}~\cite{li2017understanding} leverages the value spikes in the neuronal responses as the symptoms for fault detection. 
However, this technique has a false-positive rate of over 30\%, and program re-execution is required to restore the correct output.  
Schorn \emph{et al.}~\cite{schorn2018efficient} builds a supervised learning model to distinguish between benign and critical faults in DNNs, as well as perform error correction. 
However, their technique requires extensive FI, which is particularly time-consuming for large DNNs. For example, for a single input in the VGG16 model, there could be over 30 million values that can be corrupted by faults. Therefore, performing FIs to obtain a comprehensive training set is prohibitively expensive for large DNNs. 

 {
Hong et al.~\cite{hong2019terminal} propose an approach to mitigate memory faults in DNNs by modifying the DNN architecture, e.g., using Tanh as the activation function, which is different from \sysname's focus is on mitigating computational faults. 
{
Though they focus on memory faults (which is assumed to be protected by ECC in our work), we evaluate whether the proposed technique~\cite{hong2019terminal} is effective against computational faults as well. However, our evaluation in Fig.~\ref{fig:marland-comp} shows that this is not the case. 
}
Techniques based on Algorithm-based Fault Tolerance (ABFT) have also been proposed for detecting faults in particular layers in DNNs (e.g., {Conv layer~\cite{zhao2020algorithm,ozen2019sanity,hari2020making}; matrix multiplication~\cite{dos2018analyzing})}. 
However, these techniques do not provide protection for faults arising in other areas of the DNN, and for the fault propagating into multiple neurons. 
}


 {
Our work differs from these papers in the following ways:
\begin{itemize}[leftmargin=*]
\item 
\sysname uses value restriction in different DNN layers as a way to rectify the faulty outputs due to transient faults. 
The other papers focus on fault detection~\cite{li2017understanding,mahmoud2020hardnn,zhao2020algorithm}; they resort to black-box ML~\cite{schorn2018efficient} or redundancy~\cite{mahmoud2020hardnn} to protect the ML program from transient faults.
\item \sysname is an automated transformation that can be integrated into existing DNNs with minimal effort. In contrast, other work often require significant effort (e.g., perform expensive FI prior to deployment~\cite{schorn2018efficient,mahmoud2020hardnn}). 
\item \sysname is highly effective in fault correction with negligible overhead. We provide a comparison with representative works in Table~\ref{sec:related-work}. 
As can be seen, existing techniques in Table~\ref{sec:related-work} all suffer from either low SDC coverage or high performance overhead. In contrast, \sysname achieves an SDC coverage of over 97\% with around 0.5\% overhead, which significantly outperforms existing techniques. 
\end{itemize}
}

\begin{table}
\begin{threeparttable}
\small
\caption{Comparison of \sysname with existing techniques on protecting DNNs from hardware transient faults. }
\label{tab:comparison}
\begin{center}
\begin{tabular}{|c|c|c|}
\hline
{\bf Technique} & {\bf SDC coverage } & {\bf Overhead }\\ 
\hline
Triple Modular Redundancy & 100\%  & 200\% \\
\hline
Selective duplication~\cite{mahmoud2020hardnn} & $\sim$60\%  & 30\%  \\
\hline
Symptom-based detector~\cite{li2017understanding} \tnote{1}  & 99.5\% & 74.48\%\\ 
\hline 
ML-based error corrector~\cite{schorn2018efficient}  & 66.95\% & 0.95\% \\
\hline
Hong \emph{et al.}~\cite{hong2019terminal}\tnote{2}  & 31.54\% & 0\% \\
\hline
ABFT-based approach~\cite{zhao2020algorithm} \tnote{3} & 29.98\% & $<8\%$ \\
\hline
\textbf{\sysname (Ours)}  & \textbf{97.05\%}  & \textbf{0.53\%} \\ 
\hline
\end{tabular}

\begin{tablenotes}
        \footnotesize
        \item[1] Li \emph{et al.}~\cite{li2017understanding} did not report overhead numbers, and so we re-implemented their technique with our best effort, and measured them.
        \item[2] Results are based on the evaluation in Fig.~\ref{fig:marland-comp}. The approach in Hong \emph{et al.} does not incur any overhead as it modifies architecture itself such as replacing the activation function with another one. 
        \item[3] Zhao \emph{et al.}\cite{zhao2020algorithm} injected faults only into the convolution layer. So we estimated their overall  coverage based on the proportion of faults in the convolution layer compared to the entire network.  
\end{tablenotes} 

\end{center}
\end{threeparttable}
\end{table}


\emph{Approximate computing in ML:}
Many studies have used approximate computing techniques to lower the computation demands for DNNs~\cite{venkataramani2014axnn,judd2018proteus,hanif2018error}. Leveraging the resilience of DNNs to inexactness in the computations, these studies propose to identify those neurons that have low impact on the final output. They then either replace them with an approximate low-precision design~\cite{venkataramani2014axnn} or even remove them altogether~\cite{hanif2018error}. While our study also exploits the inherent resilience of DNNs, we restrict the ranges of values in selected DNN layers rather than find sensitive neurons in the DNNs. 

\emph{Value truncation in ML:}
Value truncation has been widely used in the ML domain for various purposes (e.g., performance~\cite{goodfellow2016deep,langford2009sparse}; robustness to outliers~\cite{wu2007robust}; privacy~\cite{ji2014differential,shokri2015privacy}). 
Gradient clipping is used to address gradient explosion during the training of DNN by rescaling the gradients to a certain range~\cite{goodfellow2016deep}. 
The loss function of the model can also be applied with truncation to improve the robustness of the learning algorithm~\cite{wu2007robust}.
Unlike these papers, we use range restriction to enable efficient (transient) fault correction.


\emph{Adversarial attack against ML:}
{
While our work studies how to rectify output corruption due to hardware transient faults, such failures can also be induced by manipulating the inputs to the ML models via adversarial attacks~\cite{papernot2018sok}.
Our fault model (Section~\ref{sec:fault-model}) assumes that the inputs to the models are uncorrupted, and hence we do not consider adversarial attacks. 
}

%% file: sections/conclusion.tex
In this work, we propose 
\sysname, an automated fault correction technique that selectively restricts the ranges of values in particular DNN layers, in order to dampen the large deviations typically caused by transient faults leading to Silent Data Corruptions (SDCs). The reduced deviations can be tolerated by the inherent resilience of DNNs, without causing SDCs. \sysname is an automated transformation that can be integrated into existing DNNs with minimal programmer effort. 

We evaluate \sysname on 8 popular DNN models, 6 of which are classifiers, and two are steering models used in the AV domain. Our evaluation demonstrates that \sysname can reduce the SDC rates of the classifier DNNs from $14.92\%$ to $0.44\%$, and those of the steering models from $23.76\%$ to $2.49\%$, without degrading the accuracy of any of the original models, and incurring negligible memory and performance overheads.

As future work, we plan to (1) {explore the combination of \sysname with existing techniques such as ML-based approaches, to reduce the remaining margin of SDCs,} and (2) extend \sysname to other ML frameworks than TensorFlow.


%% file: main.bbl
\begin{thebibliography}{10}
\providecommand{\url}[1]{#1}
\csname url@samestyle\endcsname
\providecommand{\newblock}{\relax}
\providecommand{\bibinfo}[2]{#2}
\providecommand{\BIBentrySTDinterwordspacing}{\spaceskip=0pt\relax}
\providecommand{\BIBentryALTinterwordstretchfactor}{4}
\providecommand{\BIBentryALTinterwordspacing}{\spaceskip=\fontdimen2\font plus
\BIBentryALTinterwordstretchfactor\fontdimen3\font minus
  \fontdimen4\font\relax}
\providecommand{\BIBforeignlanguage}[2]{{%
\expandafter\ifx\csname l@#1\endcsname\relax
\typeout{** WARNING: IEEEtran.bst: No hyphenation pattern has been}%
\typeout{** loaded for the language `#1'. Using the pattern for}%
\typeout{** the default language instead.}%
\else
\language=\csname l@#1\endcsname
\fi
#2}}
\providecommand{\BIBdecl}{\relax}
\BIBdecl

\bibitem{brown2020language}
T.~B. Brown, B.~Mann, N.~Ryder, M.~Subbiah, J.~Kaplan, P.~Dhariwal,
  A.~Neelakantan, P.~Shyam, G.~Sastry, A.~Askell \emph{et~al.}, ``Language
  models are few-shot learners,'' \emph{arXiv preprint arXiv:2005.14165}, 2020.

\bibitem{he2016deep}
K.~He, X.~Zhang, S.~Ren, and J.~Sun, ``Deep residual learning for image
  recognition,'' in \emph{Proceedings of the IEEE conference on computer vision
  and pattern recognition}, 2016, pp. 770--778.

\bibitem{krizhevsky2012imagenet}
A.~Krizhevsky, I.~Sutskever, and G.~E. Hinton, ``Imagenet classification with
  deep convolutional neural networks,'' in \emph{Advances in neural information
  processing systems}, 2012, pp. 1097--1105.

\bibitem{julian2016policy}
K.~D. Julian, J.~Lopez, J.~S. Brush, M.~P. Owen, and M.~J. Kochenderfer,
  ``Policy compression for aircraft collision avoidance systems,'' in
  \emph{2016 IEEE/AIAA 35th Digital Avionics Systems Conference (DASC)}.\hskip
  1em plus 0.5em minus 0.4em\relax IEEE, 2016, pp. 1--10.

\bibitem{xiong2017robust}
Z.~Xiong, M.~K. Stiles, and J.~Zhao, ``Robust ecg signal classification for
  detection of atrial fibrillation using a novel neural network,'' in
  \emph{2017 Computing in Cardiology (CinC)}.\hskip 1em plus 0.5em minus
  0.4em\relax IEEE, 2017, pp. 1--4.

\bibitem{rajpurkar2017cardiologist}
P.~Rajpurkar, A.~Y. Hannun, M.~Haghpanahi, C.~Bourn, and A.~Y. Ng,
  ``Cardiologist-level arrhythmia detection with convolutional neural
  networks,'' \emph{arXiv preprint arXiv:1707.01836}, 2017.

\bibitem{esteva2017dermatologist}
A.~Esteva, B.~Kuprel, R.~A. Novoa, J.~Ko, S.~M. Swetter, H.~M. Blau, and
  S.~Thrun, ``Dermatologist-level classification of skin cancer with deep
  neural networks,'' \emph{Nature}, vol. 542, no. 7639, p. 115, 2017.

\bibitem{bojarski2016end}
M.~Bojarski, D.~Del~Testa, D.~Dworakowski, B.~Firner, B.~Flepp, P.~Goyal, L.~D.
  Jackel, M.~Monfort, U.~Muller, J.~Zhang \emph{et~al.}, ``End to end learning
  for self-driving cars,'' \emph{arXiv preprint arXiv:1604.07316}, 2016.

\bibitem{snir2014addressing}
M.~Snir, R.~W. Wisniewski, J.~A. Abraham, S.~V. Adve, S.~Bagchi, P.~Balaji,
  J.~Belak, P.~Bose, F.~Cappello, B.~Carlson \emph{et~al.}, ``Addressing
  failures in exascale computing,'' \emph{The International Journal of High
  Performance Computing Applications}, vol.~28, no.~2, pp. 129--173, 2014.

\bibitem{oliveira2017experimental}
D.~Oliveira, L.~Pilla, N.~DeBardeleben, S.~Blanchard, H.~Quinn, I.~Koren,
  P.~Navaux, and P.~Rech, ``Experimental and analytical study of xeon phi
  reliability,'' in \emph{Proceedings of the International Conference for High
  Performance Computing, Networking, Storage and Analysis}, 2017, pp. 1--12.

\bibitem{gomez2014gpgpus}
L.~B. Gomez, F.~Cappello, L.~Carro, N.~DeBardeleben, B.~Fang, S.~Gurumurthi,
  K.~Pattabiraman, P.~Rech, and M.~S. Reorda, ``Gpgpus: how to combine high
  computational power with high reliability,'' in \emph{2014 Design, Automation
  \& Test in Europe Conference \& Exhibition (DATE)}.\hskip 1em plus 0.5em
  minus 0.4em\relax IEEE, 2014, pp. 1--9.

\bibitem{li2017understanding}
G.~Li, S.~K.~S. Hari, M.~Sullivan, T.~Tsai, K.~Pattabiraman, J.~Emer, and S.~W.
  Keckler, ``Understanding error propagation in deep learning neural network
  (dnn) accelerators and applications,'' in \emph{Proceedings of the
  International Conference for High Performance Computing, Networking, Storage
  and Analysis}.\hskip 1em plus 0.5em minus 0.4em\relax ACM, 2017, p.~8.

\bibitem{chen2019}
Z.~Chen, G.~Li, K.~Pattabiraman, and N.~DeBardeleben, ``Binfi: An efficient
  fault injector for safety-critical machine learning systems,'' in
  \emph{Proceedings of the International Conference for High Performance
  Computing, Networking, Storage and Analysis}, 2019, pp. 1--23.

\bibitem{schorn2018efficient}
C.~Schorn, A.~Guntoro, and G.~Ascheid, ``Efficient on-line error detection and
  mitigation for deep neural network accelerators,'' in \emph{International
  Conference on Computer Safety, Reliability, and Security}.\hskip 1em plus
  0.5em minus 0.4em\relax Springer, 2018, pp. 205--219.

\bibitem{hoang2019ft}
L.-H. Hoang, M.~A. Hanif, and M.~Shafique, ``Ft-clipact: Resilience analysis of
  deep neural networks and improving their fault tolerance using clipped
  activation,'' \emph{arXiv preprint arXiv:1912.00941}, 2019.

\bibitem{mahmoud2020hardnn}
A.~Mahmoud, S.~K.~S. Hari, C.~W. Fletcher, S.~V. Adve, C.~Sakr, N.~Shanbhag,
  P.~Molchanov, M.~B. Sullivan, T.~Tsai, and S.~W. Keckler, ``Hardnn: Feature
  map vulnerability evaluation in cnns,'' \emph{arXiv preprint
  arXiv:2002.09786}, 2020.

\bibitem{zhao2020algorithm}
K.~Zhao, S.~Di, S.~Li, X.~Liang, Y.~Zhai, J.~Chen, K.~Ouyang, F.~Cappello, and
  Z.~Chen, ``Algorithm-based fault tolerance for convolutional neural
  networks,'' \emph{arXiv preprint arXiv:2003.12203}, 2020.

\bibitem{ozen2019sanity}
E.~Ozen and A.~Orailoglu, ``Sanity-check: Boosting the reliability of
  safety-critical deep neural network applications,'' in \emph{2019 IEEE 28th
  Asian Test Symposium (ATS)}.\hskip 1em plus 0.5em minus 0.4em\relax IEEE,
  2019, pp. 7--75.

\bibitem{hong2019terminal}
S.~Hong, P.~Frigo, Y.~Kaya, C.~Giuffrida, and T.~Dumitra{\c{s}}, ``Terminal
  brain damage: Exposing the graceless degradation in deep neural networks
  under hardware fault attacks,'' \emph{arXiv preprint arXiv:1906.01017}, 2019.

\bibitem{AV-dataGenRate1}
\BIBentryALTinterwordspacing
``Training ai for self-driving vehicles: the challenge of scale.'' [Online].
  Available:
  \url{https://devblogs.nvidia.com/training-self-driving-vehicles-challenge-scale/}
\BIBentrySTDinterwordspacing

\bibitem{av-fitRate}
\BIBentryALTinterwordspacing
``Autonomous car - a new driver for resilient computing and design-for-test.''
  [Online]. Available:
  \url{https://nepp.nasa.gov/workshops/etw2016/talks/15WED/20160615-0930-Autonomous_Saxena-Nirmal-Saxena-Rec2016Jun16-nasaNEPP.pdf}
\BIBentrySTDinterwordspacing

\bibitem{abadi2016tensorflow}
M.~Abadi, P.~Barham, J.~Chen, Z.~Chen, A.~Davis, J.~Dean, M.~Devin,
  S.~Ghemawat, G.~Irving, M.~Isard \emph{et~al.}, ``Tensorflow: A system for
  large-scale machine learning,'' in \emph{12th $\{$USENIX$\}$ Symposium on
  Operating Systems Design and Implementation ($\{$OSDI$\}$ 16)}, 2016, pp.
  265--283.

\bibitem{tf-popularity}
\BIBentryALTinterwordspacing
``Tensorflow popularity.'' [Online]. Available:
  \url{https://towardsdatascience.com/deep-learning-framework-power-scores-2018-23607ddf297a}
\BIBentrySTDinterwordspacing

\bibitem{goodfellow2016deep}
I.~Goodfellow, Y.~Bengio, A.~Courville, and Y.~Bengio, \emph{Deep
  learning}.\hskip 1em plus 0.5em minus 0.4em\relax MIT press, 2016.

\bibitem{langford2009sparse}
J.~Langford, L.~Li, and T.~Zhang, ``Sparse online learning via truncated
  gradient,'' \emph{Journal of Machine Learning Research}, vol.~10, no. Mar,
  pp. 777--801, 2009.

\bibitem{wu2007robust}
Y.~Wu and Y.~Liu, ``Robust truncated hinge loss support vector machines,''
  \emph{Journal of the American Statistical Association}, vol. 102, no. 479,
  pp. 974--983, 2007.

\bibitem{ji2014differential}
Z.~Ji, Z.~C. Lipton, and C.~Elkan, ``Differential privacy and machine learning:
  a survey and review,'' \emph{arXiv preprint arXiv:1412.7584}, 2014.

\bibitem{shokri2015privacy}
R.~Shokri and V.~Shmatikov, ``Privacy-preserving deep learning,'' in
  \emph{Proceedings of the 22nd ACM SIGSAC conference on computer and
  communications security}, 2015, pp. 1310--1321.

\bibitem{AV-dataGenRate2}
\BIBentryALTinterwordspacing
``Autonomous and adas test cars produce over 11 tb of data per day.'' [Online].
  Available:
  \url{https://www.tuxera.com/blog/autonomous-and-adas-test-cars-produce-over-11-tb-of-data-per-day/}
\BIBentrySTDinterwordspacing

\bibitem{iso26262}
\BIBentryALTinterwordspacing
``Functional safety methodologies for automotive applications.'' [Online].
  Available:
  \url{https://www.cadence.com/content/dam/cadence-www/global/en_US/documents/solutions/automotive-functional-safety-wp.pdf}
\BIBentrySTDinterwordspacing

\bibitem{pei2017deepxplore}
K.~Pei, Y.~Cao, J.~Yang, and S.~Jana, ``Deepxplore: Automated whitebox testing
  of deep learning systems,'' in \emph{proceedings of the 26th Symposium on
  Operating Systems Principles}.\hskip 1em plus 0.5em minus 0.4em\relax ACM,
  2017, pp. 1--18.

\bibitem{tian2018deeptest}
Y.~Tian, K.~Pei, S.~Jana, and B.~Ray, ``Deeptest: Automated testing of
  deep-neural-network-driven autonomous cars,'' in \emph{Proceedings of the
  40th international conference on software engineering}.\hskip 1em plus 0.5em
  minus 0.4em\relax ACM, 2018, pp. 303--314.

\bibitem{rubaiyat2018experimental}
A.~H.~M. Rubaiyat, Y.~Qin, and H.~Alemzadeh, ``Experimental resilience
  assessment of an open-source driving agent,'' \emph{arXiv preprint
  arXiv:1807.06172}, 2018.

\bibitem{yolo}
\BIBentryALTinterwordspacing
``Yolo v5.'' [Online]. Available:
  \url{https://github.com/ultralytics/yolov5.git}
\BIBentrySTDinterwordspacing

\bibitem{banerjee2018hands}
S.~S. Banerjee, S.~Jha, J.~Cyriac, Z.~T. Kalbarczyk, and R.~K. Iyer, ``Hands
  off the wheel in autonomous vehicles?: A systems perspective on over a
  million miles of field data,'' in \emph{2018 48th Annual IEEE/IFIP
  International Conference on Dependable Systems and Networks (DSN)}.\hskip 1em
  plus 0.5em minus 0.4em\relax IEEE, 2018, pp. 586--597.

\bibitem{reagen2018ares}
B.~Reagen, U.~Gupta, L.~Pentecost, P.~Whatmough, S.~K. Lee, N.~Mulholland,
  D.~Brooks, and G.-Y. Wei, ``Ares: A framework for quantifying the resilience
  of deep neural networks,'' in \emph{2018 55th ACM/ESDA/IEEE Design Automation
  Conference (DAC)}.\hskip 1em plus 0.5em minus 0.4em\relax IEEE, 2018, pp.
  1--6.

\bibitem{NIPS2019_8810}
H.~Guan, L.~Ning, Z.~Lin, X.~Shen, H.~Zhou, and S.-H. Lim, ``In-place
  zero-space memory protection for cnn,'' \emph{arXiv preprint
  arXiv:1910.14479}, 2019.

\bibitem{li2018modeling}
G.~Li, K.~Pattabiraman, S.~K.~S. Hari, M.~Sullivan, and T.~Tsai, ``Modeling
  soft-error propagation in programs,'' in \emph{2018 48th Annual IEEE/IFIP
  International Conference on Dependable Systems and Networks (DSN)}.\hskip 1em
  plus 0.5em minus 0.4em\relax IEEE, 2018, pp. 27--38.

\bibitem{ashraf2015understanding}
R.~A. Ashraf, R.~Gioiosa, G.~Kestor, R.~F. DeMara, C.-Y. Cher, and P.~Bose,
  ``Understanding the propagation of transient errors in hpc applications,'' in
  \emph{SC'15: Proceedings of the International Conference for High Performance
  Computing, Networking, Storage and Analysis}.\hskip 1em plus 0.5em minus
  0.4em\relax IEEE, 2015, pp. 1--12.

\bibitem{georgakoudis2017refine}
G.~Georgakoudis, I.~Laguna, D.~S. Nikolopoulos, and M.~Schulz, ``Refine:
  Realistic fault injection via compiler-based instrumentation for accuracy,
  portability and speed,'' in \emph{Proceedings of the International Conference
  for High Performance Computing, Networking, Storage and Analysis}.\hskip 1em
  plus 0.5em minus 0.4em\relax ACM, 2017, p.~29.

\bibitem{fang2014gpu}
B.~Fang, K.~Pattabiraman, M.~Ripeanu, and S.~Gurumurthi, ``Gpu-qin: A
  methodology for evaluating the error resilience of gpgpu applications,'' in
  \emph{2014 IEEE International Symposium on Performance Analysis of Systems
  and Software (ISPASS)}.\hskip 1em plus 0.5em minus 0.4em\relax IEEE, 2014,
  pp. 221--230.

\bibitem{wei2014quantifying}
J.~Wei, A.~Thomas, G.~Li, and K.~Pattabiraman, ``Quantifying the accuracy of
  high-level fault injection techniques for hardware faults,'' in \emph{2014
  44th Annual IEEE/IFIP International Conference on Dependable Systems and
  Networks}.\hskip 1em plus 0.5em minus 0.4em\relax IEEE, 2014, pp. 375--382.

\bibitem{Chang2018}
C.-K. Chang, S.~Lym, N.~Kelly, M.~B. Sullivan, and M.~Erez, ``Evaluating and
  accelerating high-fidelity error injection for hpc,'' in \emph{Proceedings of
  the International Conference for High Performance Computing, Networking,
  Storage, and Analysis}, ser. SC '18, 2018, pp. 45:1--45:13.

\bibitem{sangchoolie2017one}
B.~Sangchoolie, K.~Pattabiraman, and J.~Karlsson, ``One bit is (not) enough: An
  empirical study of the impact of single and multiple bit-flip errors,'' in
  \emph{2017 47th Annual IEEE/IFIP International Conference on Dependable
  Systems and Networks (DSN)}.\hskip 1em plus 0.5em minus 0.4em\relax IEEE,
  2017, pp. 97--108.

\bibitem{chang2018evaluating}
C.-K. Chang, S.~Lym, N.~Kelly, M.~B. Sullivan, and M.~Erez, ``Evaluating and
  accelerating high-fidelity error injection for hpc,'' in \emph{Proceedings of
  the International Conference for High Performance Computing, Networking,
  Storage, and Analysis}.\hskip 1em plus 0.5em minus 0.4em\relax IEEE Press,
  2018, p.~45.

\bibitem{venkataramani2014axnn}
S.~Venkataramani, A.~Ranjan, K.~Roy, and A.~Raghunathan, ``Axnn:
  Energy-efficient neuromorphic systems using approximate computing,'' in
  \emph{2014 IEEE/ACM International Symposium on Low Power Electronics and
  Design (ISLPED)}.\hskip 1em plus 0.5em minus 0.4em\relax IEEE, 2014, pp.
  27--32.

\bibitem{hari2012relyzer}
S.~K.~S. Hari, S.~V. Adve, H.~Naeimi, and P.~Ramachandran, ``Relyzer:
  Exploiting application-level fault equivalence to analyze application
  resiliency to transient faults,'' in \emph{ACM SIGPLAN Notices}, vol.~47,
  no.~4.\hskip 1em plus 0.5em minus 0.4em\relax ACM, 2012, pp. 123--134.

\bibitem{li2018tensorfi}
G.~Li, K.~Pattabiraman, and N.~DeBardeleben, ``Tensorfi: A configurable fault
  injector for tensorflow applications,'' in \emph{2018 IEEE International
  Symposium on Software Reliability Engineering Workshops (ISSREW)}.\hskip 1em
  plus 0.5em minus 0.4em\relax IEEE, 2018, pp. 313--320.

\bibitem{commaai}
\BIBentryALTinterwordspacing
``comma.ai's steering model.'' [Online]. Available:
  \url{https://github.com/commaai/research}
\BIBentrySTDinterwordspacing

\bibitem{dave-test}
\BIBentryALTinterwordspacing
``On-road tests for nvidia dave system.'' [Online]. Available:
  \url{https://devblogs.nvidia.com/deep-learning-self-driving-cars/}
\BIBentrySTDinterwordspacing

\bibitem{drivingDataset}
\BIBentryALTinterwordspacing
``Driving dataset.'' [Online]. Available:
  \url{https://github.com/SullyChen/driving-datasets}
\BIBentrySTDinterwordspacing

\bibitem{du2017self}
S.~Du, H.~Guo, and A.~Simpson, ``Self-driving car steering angle prediction
  based on image recognition,'' \emph{Department of Computer Science, Stanford
  University, Tech. Rep. CS231-626}, 2017.

\bibitem{autopilot-git}
\BIBentryALTinterwordspacing
``Tensorflow implementation of nvidia dave system.'' [Online]. Available:
  \url{https://github.com/SullyChen/Autopilot-TensorFlow}
\BIBentrySTDinterwordspacing

\bibitem{judd2018proteus}
P.~Judd, J.~Albericio, T.~Hetherington, T.~Aamodt, N.~E. Jerger, R.~Urtasun,
  and A.~Moshovos, ``Proteus: Exploiting precision variability in deep neural
  networks,'' \emph{Parallel Computing}, 2018.

\bibitem{tang2018experimental}
R.~Tang, W.~Wang, Z.~Tu, and J.~Lin, ``An experimental analysis of the power
  consumption of convolutional neural networks for keyword spotting,'' in
  \emph{2018 IEEE International Conference on Acoustics, Speech and Signal
  Processing (ICASSP)}.\hskip 1em plus 0.5em minus 0.4em\relax IEEE, 2018, pp.
  5479--5483.

\bibitem{han2015learning}
S.~Han, J.~Pool, J.~Tran, and W.~Dally, ``Learning both weights and connections
  for efficient neural network,'' in \emph{Advances in neural information
  processing systems}, 2015, pp. 1135--1143.

\bibitem{sehgal2019guidelines}
A.~Sehgal and N.~Kehtarnavaz, ``Guidelines and benchmarks for deployment of
  deep learning models on smartphones as real-time apps,'' \emph{Machine
  Learning and Knowledge Extraction}, vol.~1, no.~1, pp. 450--465, 2019.

\bibitem{yang2020practical}
L.~Yang, B.~Nie, A.~Jog, and E.~Smirni, ``Practical resilience analysis of
  gpgpu applications in the presence of single- and multi-bit faults,''
  \emph{IEEE Transactions on Computers}, vol.~70, no.~01, pp. 30--44, jan 2021.

\bibitem{reagen2016minerva}
B.~Reagen, P.~Whatmough, R.~Adolf, S.~Rama, H.~Lee, S.~K. Lee, J.~M.
  Hern{\'a}ndez-Lobato, G.-Y. Wei, and D.~Brooks, ``Minerva: Enabling
  low-power, highly-accurate deep neural network accelerators,'' in
  \emph{ACM/IEEE 43rd Annual International Symposium on Computer Architecture
  (ISCA)}.\hskip 1em plus 0.5em minus 0.4em\relax IEEE, 2016.

\bibitem{dos2019reliability}
F.~F. dos Santos, C.~Lunardi, D.~Oliveira, F.~Libano, and P.~Rech,
  ``Reliability evaluation of mixed-precision architectures,'' in \emph{2019
  IEEE International Symposium on High Performance Computer Architecture
  (HPCA)}.\hskip 1em plus 0.5em minus 0.4em\relax IEEE, 2019, pp. 238--249.

\bibitem{hari2020making}
S.~K.~S. Hari, M.~B. Sullivan, T.~Tsai, and S.~W. Keckler, ``Making
  convolutions resilient via algorithm-based error detection techniques,''
  \emph{arXiv preprint arXiv:2006.04984}, 2020.

\bibitem{dos2018analyzing}
F.~F. dos Santos, P.~F. Pimenta, C.~Lunardi, L.~Draghetti, L.~Carro, D.~Kaeli,
  and P.~Rech, ``Analyzing and increasing the reliability of convolutional
  neural networks on gpus,'' \emph{IEEE Transactions on Reliability}, vol.~68,
  no.~2, pp. 663--677, 2018.

\bibitem{hanif2018error}
M.~A. Hanif, R.~Hafiz, and M.~Shafique, ``Error resilience analysis for
  systematically employing approximate computing in convolutional neural
  networks,'' in \emph{2018 Design, Automation \& Test in Europe Conference \&
  Exhibition (DATE)}.\hskip 1em plus 0.5em minus 0.4em\relax IEEE, 2018.

\bibitem{papernot2018sok}
N.~Papernot, P.~McDaniel, A.~Sinha, and M.~P. Wellman, ``Sok: Security and
  privacy in machine learning,'' in \emph{2018 IEEE European Symposium on
  Security and Privacy (EuroS\&P)}.\hskip 1em plus 0.5em minus 0.4em\relax
  IEEE, 2018, pp. 399--414.

\end{thebibliography}
